\documentclass[journal,final]{IEEEtran}

\usepackage{cite}
\usepackage{amsfonts,times,amsmath,amssymb,amsthm}
\usepackage{algorithmic}
\usepackage{textcomp}
\usepackage{bm}
\usepackage{float}
\usepackage{lipsum}
\usepackage{mathrsfs}
\usepackage{makecell}
\usepackage{slashbox}
\usepackage{dblfloatfix}
\usepackage{commath}
\usepackage{amstext}
\usepackage{mathtools}
\usepackage{cuted}
\usepackage{algorithm}
\usepackage{algorithmic}
\usepackage{hyperref}
\usepackage{graphics}
\usepackage{epsfig}
\usepackage{epstopdf}

\usepackage{colortbl}             
\usepackage[table,xcdraw]{xcolor} 
\usepackage{booktabs}             
\usepackage{graphicx}             
\usepackage{subcaption}           
\usepackage{multirow}

\theoremstyle{plain}

\theoremstyle{definition}

\theoremstyle{remark}

\graphicspath{{images/}}

\begin{document}
\title{Drone Carrier: An Integrated Unmanned Surface Vehicle for Autonomous Inspection and Intervention in GNSS-Denied Maritime Environment}
\author{Yihao Dong,	Muhayyu~Ud~Din, Francesco Lagala, Hailiang Kuang, Jianjun Sun, Siyuan Yang, Irfan Hussain and Shaoming He\IEEEauthorrefmark{1}

\thanks{This work is supported by the National Key Research and Development Program under Grant No. 2022YFE0204400, the Beijing Science and Technology Project under Grant No. Z221100002722008, and the Khalifa University under Award No. RC1-2018-KUCARS-8474000136, CIRA-2021-085, MBZIRC-8434000194, KU-BIT-Joint-Lab-8434000534.}
\thanks{Yihao Dong, Muhayy Ud Din and Irfan Hussain are with Khalifa University Center of Autonomous Robotic System (KUCARS), 127788 Abu Dhabi, UAE.}
\thanks{Francesco Lagala is with the Institute for Marine Engineering, National Research Council, 139-00128 Rome, Italy.}
\thanks{Hailiang Kuang, Jianjun Sun, Siyuan Yang and Shaoming He are with the School of Aerospace Engineering, Beijing Institute of Technology, Beijing, 100081 China.} 
\thanks{\IEEEauthorrefmark{1}Corresponding Author. Email: $\texttt{shaoming.he@bit.edu.cn}$}
}

\maketitle

\begin{abstract}
This paper introduces an innovative drone carrier concept that is applied in maritime port security or offshore rescue. This system works with a heterogeneous system consisting of multiple Unmanned Aerial Vehicles (UAVs) and Unmanned Surface Vehicles (USVs) to perform inspection and intervention tasks in GNSS-denied or interrupted environments. The carrier, an electric catamaran measuring 4m by 7m, features a 4m by 6m deck supporting automated takeoff and landing for four DJI M300 drones, along with a 10kg-payload manipulator operable in up to level 3 sea conditions. Utilizing an offshore gimbal camera for navigation, the carrier can autonomously navigate, approach and dock with non-cooperative vessels, guided by an onboard camera, LiDAR, and Doppler Velocity Log (DVL) over a 3 km$^2$ area. UAVs equipped with onboard Ultra-Wideband (UWB) technology execute mapping, detection, and manipulation tasks using a versatile gripper designed for wet, saline conditions. Additionally, two UAVs can coordinate to transport large objects to the manipulator or interact directly with them. These procedures are fully automated and were successfully demonstrated at the Mohammed Bin Zayed International Robotic Competition (MBZIRC2024), where the drone carrier equipped with four UAVS and one manipulator, automatically accomplished the intervention tasks in sea-level-3 (wave height 1.25m) based on the rough target information. 
\end{abstract}


\section{Introduction}

\IEEEPARstart{M}{arine} robotics has garnered significant attention in recent years due to its various applications in maritime security \cite{johnston2017marine}, environmental monitoring \cite{Villa2016}, and disaster response \cite{jorge2019survey}, rescue operations \cite{matos2013development}. Compared to crewed vessels, marine robots offer significant advantages for executing repetitive tasks over extended durations and across large spatial scales, as well as for undertaking hazardous missions in poorly characterized or unknown environments. The deployment of unmanned systems for search and intervention operations results in increased inspection efficiency, improved operational capabilities, and a reduction in risks to personnel.

The complementary capabilities of USVs and UAVs have led to their increasing deployment as heterogeneous systems for maritime inspection and intervention. UAVs extend the operational range of USVs, enabling coverage of coastal zones \cite{Lindemuth2011}, offshore energy infrastructure \cite{Collins2017}, and oil spill monitoring \cite{Vasilijevic2015}. Moreover, UAVs enhance the effectiveness of USV-based search and rescue operations \cite{Gallego2019,Mendonca2016}. Current approaches typically utilize Global Navigation Satellite System (GNSS) for localizing the heterogeneous system, which is remotely operated from a command center to execute predefined search and intervention tasks. In GNSS-denied environments, the heterogeneous system can be localized through sensor fusion, integrating inertial measurement unit (IMU) data with onboard USV sensors such as radar \cite{Han2019}, LiDAR \cite{Shen2023}, or cameras \cite{Ma2018}. Maritime intervention tasks, including marine rescues \cite{ferrao2022security} and cargo handling (loading/unloading) \cite{jofre2021implementation,brandao2022side}, present further challenges related to the trade-off between UAV payload capacity and the limited reach of onboard manipulators \cite{wang2023applications}. Furthermore, these robotic systems are vulnerable to disruptions, particularly in adverse weather conditions such as high sea states and dense fog, which can degrade data links and GNSS availability. Intentional interference from unauthorized vessels or within restricted areas also poses a threat to system integrity and can potentially lead to complete system failure.


The proposed drone carrier is specifically designed for autonomous maritime inspection and intervention operations in GNSS-denied or GNSS-challenged environments. Its modular design allows for scalable configurations, ranging from a minimal deck size of 2m $\times$ 1.5m, accommodating a single UAV, to a maximal configuration of 8m $\times$ 7m, capable of deploying 12 standard UAVs and a robotic manipulator, thereby adapting to diverse operational requirements. The USV is equipped with multiple sensors for localization and navigation, including a DVL, an IMU, LiDARs, cameras, and infrared thermal imagers. Data from these sensors are processed by an onboard computer, with only essential information transmitted to the command center to facilitate informed decision-making. The unobstructed, open deck provides ample space for UAV takeoff and landing, with QR codes assisting UAV precise landing adjustments, compensating for variations in altitude and sea state. Six UWB transceivers mounted above the deck establish a local positioning system for the UAVs, while two 2.4 GHz antennas provide data link connectivity between the drone carrier, the command center, and other carriers. An enhanced Wi-Fi network within the carrier facilitates communication between onboard sensors and the UAVs. To ensure compatibility, each UAV is equipped with an integrated system comprising an onboard computer, a downward-facing landing camera, and a UWB transceiver. Initial navigation of the overall system relies on an onshore camera, transitioning to onboard sensor-based navigation once targets are detected by the USV's onboard cameras or LiDARs. The proposed drone carrier is the only team to successfully complete inspection and intervention tasks in a sea-level-3, GNSS-denied sea environment during the MBZIRC2024 demonstrations\footnote{A video of the field test demonstration during the MBZIRC 2024 can be found at: https://www.youtube.com/watch?v=w5ciWKv-yAQ.}. The main contributions of this paper include:
\begin{enumerate}
	\item Design of a modular USV-based Drone Carrier. The carrier is equipped with integrated multi-domain sensors, intelligent drones and a manipulator, enabling it to perform investigation and intervention tasks in GNSS-denied sea environments. Its fully electric and modular design supports environmental sustainability and is adaptable to specific operational scenarios.
	
	\item Robust, integrated multifunction robotic System. This software architecture, specifically designed for inspection and intervention operations in GNSS-denied sea environments, includes multiple drones and manipulators. It features four foundational layers—perception, recognition, decision, and action—crafted to enhance the autonomy of the system, eliminating the need for human intervention.
	
	\item Comprehensive experimental validation. Extensive tests conducted in real sea environments (with a wave height of 1.5 m and a wind speed of 8 m/s) demonstrate the system’s capabilities. These tests include approaching and docking with a non-cooperative target vessel, performing intervention tasks using a manipulator, and transporting small targets via drone. These trials illustrate the system’s ability to autonomously execute inspection and intervention tasks in challenging GNSS-denied conditions.
\end{enumerate}

The document is arranged as follows: Section II  presents the related work with a brief overview of the current USV-UAV heterogeneous system. In Section III, the drone carrier' concept and hardware architecture are presented, followed by key algorithms provided in Section IV. Section V depicts the drone carrier's real-world experiments and data with different conditions in real scenarios. Finally, some final remarks and conclusions are offered.

\section{Related Work}
Autonomous inspection and intervention in the sea environment is a harsh task since multiple robotics systems are required to cooperate considering mission constraints and environmental disturbances. Thanks to the natural complementary of UAV and USV, the current solution uses a USV-UAV heterogeneous system, collaboratively conducting searching and transportation tasks \cite{young2017robot,Gallego2019,specht2024methodology}. Multiple onboard sensors on both UAV and USV monitor the system condition and obey the decision from the remote command center. Early implementations of UAV-USV systems for coastal inspection, as described in \cite{Lindemuth2011}, emphasized the need for high levels of vehicle autonomy for effective collaborative platforms and swarm deployments. Subsequent research has explored various applications, including harmful algal bloom mitigation using UAV-based detection and USV-based removal \cite{Jung2017}, inspection of offshore energy infrastructure using UAVs and visual tracking of the USV \cite{Collins2017}, and disaster management, where UAVs tracked USVs to assess littoral structural damage following Hurricane Wilma \cite{Murphy2008}. UAV-USV systems have also been employed in oil spill monitoring \cite{Vasilijevic2015} and search and rescue operations \cite{Gallego2019,Mendonca2016}. Vision-based USV navigation aided by UAVs has been investigated \cite{Xiao2017}, and the critical process of UAV landing on a USV has been studied through numerical simulations using relative motion modeling \cite{Sun2020, Huang2018}.

Achieving close coordination within these heterogeneous UAV-USV systems requires autonomous UAV takeoff and landing on the USV platform. Unlike landing on static ground platforms, autonomous landing on a moving and oscillating USV necessitates robust state estimation, dynamic motion coupling between the UAV and USV, and stable UAV flight control. Huang et al. \cite{Huang2018} implemented Adaptive Sliding Mode Control, demonstrating robustness in landing tasks despite environmental disturbances and uncertainties. Shao et al. \cite{Shao2019} developed a cooperative platform for secure UAV landing on a vessel deck, using four ultrasonic sensors on the USV deck to guide the UAV to the landing area. Tian et al. \cite{tian2024uav} demonstrated that estimating the USV's oscillatory state and landing within a reasonable oscillation range significantly improves UAV landing precision and success rate.

Most unmanned maritime systems rely on GNSS and IMUs for localization and state estimation. However, GNSS signals are vulnerable to both intentional and unintentional interference, rendering them unreliable in certain environments, particularly in coastal waters where GNSS jamming is increasingly prevalent. Consequently, alternative navigation solutions are crucial for safe operation. Several studies have addressed localization in GNSS-denied environments. Liu et al. \cite{Liu2023} explored a visual-inertial odometry approach for USV localization, while Shen et al. \cite{Shen2023} combined LiDAR data with IMU information. However, these LiDAR and vision-based methods require distinct environmental features for effective localization, which may be lacking in open sea areas. Han et al. \cite{Han2019} proposed a radar-based USV localization method using extracted coastal landmarks. Ma et al. \cite{Ma2018} discussed a radar-based coastal image registration technique, integrating offline satellite imagery with radar data for localization. While effective, radar-based solutions are typically expensive and power-intensive, making them less suitable for energy-constrained USVs.

Table \ref{tab:usv_uav_summary} summarizes the existing USV-UAV heterogeneous systems, characterizing their operational tasks, sensors and environmental constraints.

\begin{table*}[h!]
	\centering
	\caption{Summary of the existing USV-UAV heterogeneous system}
	\resizebox{\textwidth}{!}{%
		\begin{tabular}{p{3cm}p{0.8cm}p{2.5cm}p{0.8cm}p{0.8cm}p{0.8cm}p{0.8cm}p{0.8cm}p{0.8cm}p{0.8cm}p{0.8cm}p{0.8cm}p{0.8cm}p{0.8cm}p{0.8cm}p{0.8cm}p{0.8cm}p{0.8cm}}
			\toprule
			&    &  & \multicolumn{10}{c}{\cellcolor[HTML]{EFEFEF}\textbf{USV}}   & \multicolumn{4}{c}{\cellcolor[HTML]{EFEFEF}\textbf{UAV}} &  \\ \textbf{Organization \& Reference}    & \textbf{Year} & \textbf{Main Task}& \textbf{UWB}& \textbf{Data link}       & \textbf{Land Area}       & \textbf{\begin{tabular}[c]{@{}l@{}}Com-\\ pass\end{tabular}}& \textbf{IMU}& \textbf{\begin{tabular}[c]{@{}l@{}}Li-\\ DAR\end{tabular}}  & \textbf{DVL}& \textbf{\begin{tabular}[c]{@{}l@{}}Cam-\\ era\end{tabular}} & \textbf{Rob. arm}& \textbf{UAV type} & \textbf{\begin{tabular}[c]{@{}l@{}}Num-\\ ber\end{tabular}}   & \textbf{\begin{tabular}[c]{@{}l@{}}Gri-\\ pper\end{tabular}}& \textbf{\begin{tabular}[c]{@{}l@{}}Li-\\ DAR\end{tabular}}  & \textbf{\begin{tabular}[c]{@{}l@{}}Cam-\\ era\end{tabular}} & \textbf{GNSS Denied} \\ \midrule
			\rowcolor[HTML]{EFEFEF} University of South Florida \cite{Murphy2008}  & 2008  & Inspection &   & \cellcolor[HTML]{333333} & & & \cellcolor[HTML]{333333} & & & \cellcolor[HTML]{333333} & & \begin{tabular}[c]{@{}l@{}}Helic-\\ optor\end{tabular}       & 1& & & \cellcolor[HTML]{333333} &               \\
			FCT-UNL   \cite{Mendonca2016}    & 2016  & Search and rescue&   & \cellcolor[HTML]{333333} & \cellcolor[HTML]{333333} & & \cellcolor[HTML]{333333} & & & \cellcolor[HTML]{333333} & & Hexa copter       & 1& & & \cellcolor[HTML]{333333} &               \\
			\rowcolor[HTML]{EFEFEF} University of Illinois at Urbana-Champaign \cite{young2017robot} & 2017  & Measurement for Hydrologic    &   & & & & \cellcolor[HTML]{333333} & & & \cellcolor[HTML]{333333} & & Quad  rotor& 1& & & \cellcolor[HTML]{333333} &               \\
			Texas A\&M University \cite{Xiao2017}       & 2017  & Search and Rescue&   & \cellcolor[HTML]{333333} & & & \cellcolor[HTML]{333333} & & & \cellcolor[HTML]{333333} & & Quad  rotor& 1& & & \cellcolor[HTML]{333333} &               \\
			\rowcolor[HTML]{EFEFEF} Planck Aerosystems \cite{Collins2017}
			& 2017  & Offshore Inspections &   & \cellcolor[HTML]{333333} & \cellcolor[HTML]{333333} & & & \cellcolor[HTML]{333333} & & \cellcolor[HTML]{333333} & & Quad  rotor& 1& & & \cellcolor[HTML]{333333} &               \\
			University of Alicante \cite{Gallego2019}       & 2018  & Maritime Rescue Operations    &   & \cellcolor[HTML]{333333} & & & \cellcolor[HTML]{333333} & & & \cellcolor[HTML]{333333} & & Quad  rotor& 1& & & \cellcolor[HTML]{333333} &               \\
			\rowcolor[HTML]{EFEFEF} Dalian University of Technology \cite{Shao2019}  & 2018  & Inspection and Transportation &   & \cellcolor[HTML]{333333} & \cellcolor[HTML]{333333} & & \cellcolor[HTML]{333333} & & & \cellcolor[HTML]{333333} & & Quad  rotor& 1& & & \cellcolor[HTML]{333333} &               \\
			Chinese Academy of Sciences \cite{zhang2020marine}  & 2020  & Monitoring &   & \cellcolor[HTML]{333333} & \cellcolor[HTML]{333333} & & \cellcolor[HTML]{333333} & & & \cellcolor[HTML]{333333} & \cellcolor[HTML]{333333} & Fixed Wing & 1& & & \cellcolor[HTML]{333333} &               \\
			\rowcolor[HTML]{EFEFEF} KAIST \cite{Huang2018}     & 2020  & Algal Bloom Removal  &   & \cellcolor[HTML]{333333} & & & \cellcolor[HTML]{333333} & & & \cellcolor[HTML]{333333} & & Quad  rotor& 1& & & \cellcolor[HTML]{333333} &               \\
			Anhui Polytechnic University \cite{li2022synchronized}  & 2022  & Cooperative Autonomous Landing&   & \cellcolor[HTML]{333333} & \cellcolor[HTML]{333333} & & \cellcolor[HTML]{333333} & & & \cellcolor[HTML]{333333} & & Quad  rotor& 1& & & \cellcolor[HTML]{333333} &               \\
			\rowcolor[HTML]{EFEFEF} The University of Hong Kong  \cite{wei20223u}  & 2023  & Target Hunting   &   & \cellcolor[HTML]{333333} & \cellcolor[HTML]{333333} & & \cellcolor[HTML]{333333} & & & \cellcolor[HTML]{333333} & & Quad  rotor& 1& & & \cellcolor[HTML]{333333} &               \\
			Czech Technical University \cite{novak2024towards}      & 2024  & Collaboration in Harsh Maritime Conditions &   & \cellcolor[HTML]{333333} & \cellcolor[HTML]{333333} & & \cellcolor[HTML]{333333} & & & \cellcolor[HTML]{333333} & & Quad  rotor& 1& & & \cellcolor[HTML]{333333} &               \\
			\rowcolor[HTML]{EFEFEF} Gdynia Maritime University \cite{specht2024methodology}  & 2024  & Coastal Zone Measurements     &   & \cellcolor[HTML]{333333} & & & \cellcolor[HTML]{333333} & \cellcolor[HTML]{333333} & & \cellcolor[HTML]{333333} & & Hexa copter       & 1& & & \cellcolor[HTML]{333333} &               \\
			Harbin Engineering University \cite{tian2024uav}& 2024  & Cooperative Autonomous Landing&   \cellcolor[HTML]{333333} & \cellcolor[HTML]{333333} & \cellcolor[HTML]{333333} & \cellcolor[HTML]{333333} & \cellcolor[HTML]{333333} & & & \cellcolor[HTML]{333333} & & Quad  rotor& 1& & & \cellcolor[HTML]{333333} &               \\
			\rowcolor[HTML]{EFEFEF} Czech Technical University \cite{novak2024collaborative}      & 2024  & Collaborative Object Manipulation &   & \cellcolor[HTML]{333333} & \cellcolor[HTML]{333333} & & \cellcolor[HTML]{333333} & & & \cellcolor[HTML]{333333} & & Quad  rotor& 1& & & \cellcolor[HTML]{333333} &               \\
			Ours  & 2024  & Inspection and Transportation &  \cellcolor[HTML]{333333} & \cellcolor[HTML]{333333} & \cellcolor[HTML]{333333} & \cellcolor[HTML]{333333} & \cellcolor[HTML]{333333} & \cellcolor[HTML]{333333} & \cellcolor[HTML]{333333} & \cellcolor[HTML]{333333} & \cellcolor[HTML]{333333} & Quad  rotor& 4& \cellcolor[HTML]{333333} & \cellcolor[HTML]{333333} & \cellcolor[HTML]{333333} & \cellcolor[HTML]{333333}              \\ \bottomrule
		\end{tabular}
	}
	\label{tab:usv_uav_summary}
\end{table*}

\section{Hardware Description}

\subsection{Overall System}
\begin{figure*} [h]
	\centering
	\includegraphics[width=\textwidth]{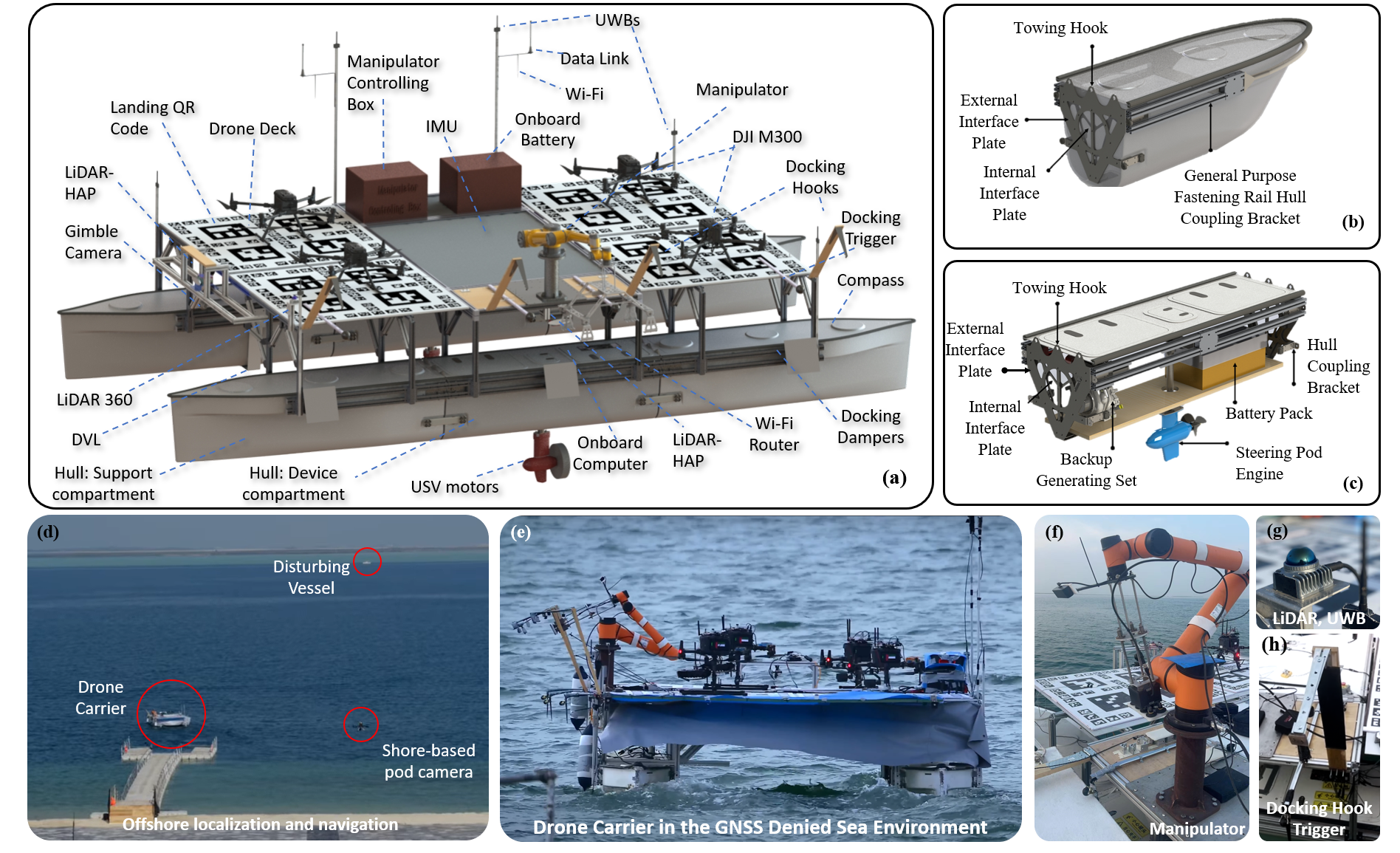}
	\caption{The proposed drone carrier concepts. (a): CAD design of the integrated constitution, (b): support, (c): device compartment, (d): Onshore-aided navigation, (e): the drone carrier in the sea, (f): manipulator, (g): LiDAR and UWB, and (h): docking hook and trigger.}
	\label{fig:System}
\end{figure*}

Inspection and intervention in the GNSS-denied marine environment require a robust platform that conducts multiple tasks in facing environmental uncertainty. The system is designed to operate in an open ocean environment, relying on approximate target location information to autonomously search, approach, and dock with a target. Once docked, the system issues takeoff commands to drones and continuously provides target location information. After the drones complete their search and transport tasks, the system guides them to land at designated locations and subsequently return to the base. To accomplish these tasks, the drone carrier system is required to finish: searching, approaching and docking, single-drone transport, and collaborative transport. The hardware subsystems include the USV, UAV and robotic arm system. One typical configuration of the proposed drone carrier consists of a 4m $\times$ 7m drone deck, carrying 4 DJI-M300 drones and 1 manipulator as illustrated in Fig. \ref{fig:System}. The following assumptions and constraints are considered for our drone carrier operations:
\begin{enumerate}
	\item The operational environment is not higher than sea state 3, with wave heights not exceeding 1.5m and atmospheric visibility greater than 2 km.
	\item The USV is able to access to prior information about the target from the onshore-aided navigation, including its approximate location with an accuracy of 50m and a limited number of target photographs. 
	\item The USV is capable of docking and attaching only onto target vessels no more than twice its size and rely on the presence of well-defined edges on the target vessel for its latching mechanism to secure hard holding. 
\end{enumerate}

\subsection{USV System}
The USV system, serving as the core component of the drone carrier, features a modularized catamaran design integrated with onboard sensors. The catamaran's configuration can be adjusted based on the required search area, operational scope, and the number of UAVs it needs to carry. All onboard sensors are housed separately from the catamaran structure, yet they effectively control the speed and orientation of the dual motors.

\begin{figure*}[ht]
	\centering
	\begin{subfigure}[t]{0.4\textwidth}
		\centering
		\includegraphics[height=8cm]{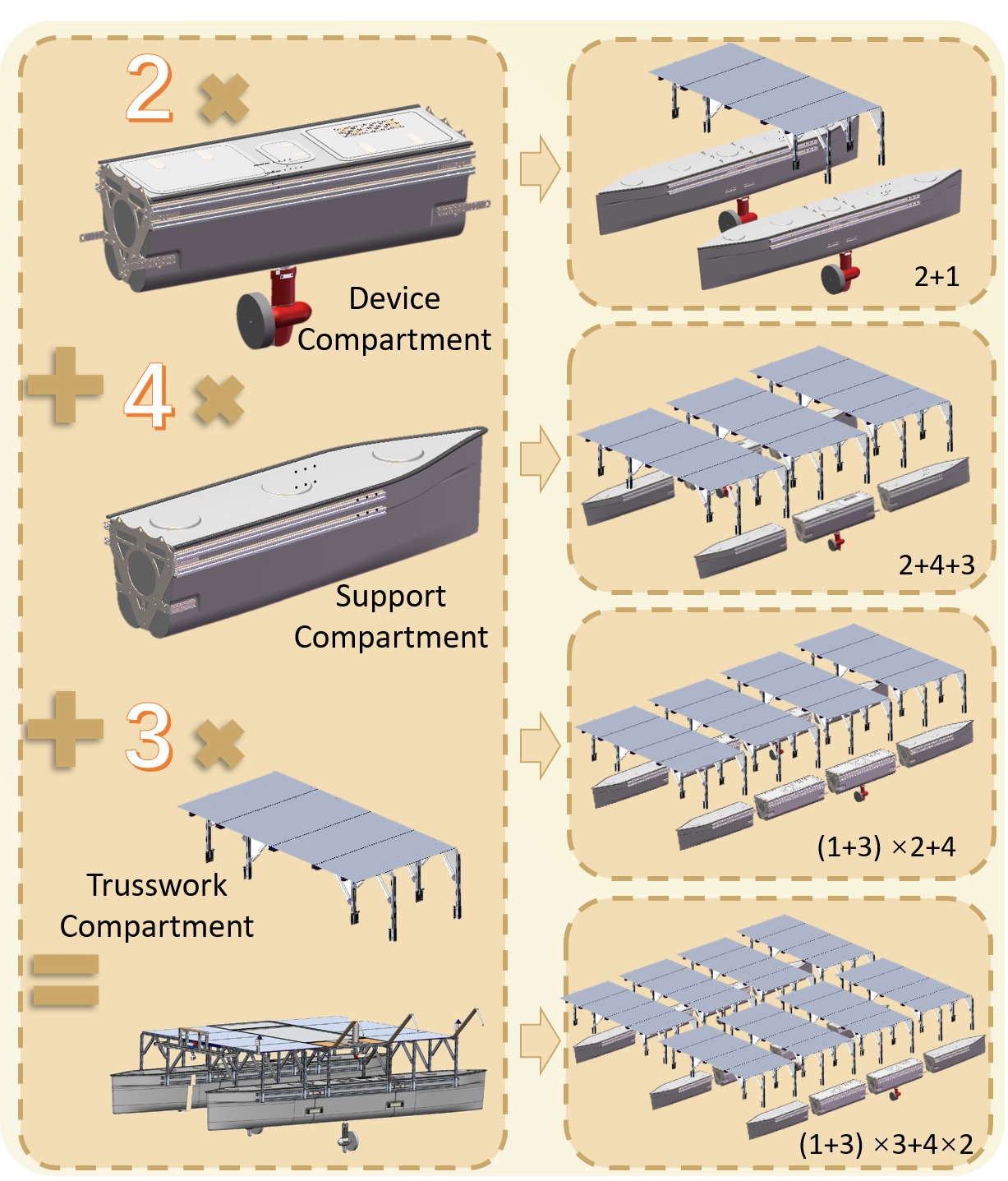} 
		\caption{Flexible assembly of unmanned ship components for diverse configurations.}
		\label{fig:sub1}
	\end{subfigure}
	\hfill
	\begin{subfigure}[t]{0.59\textwidth}
		\centering
		\includegraphics[height=8cm]{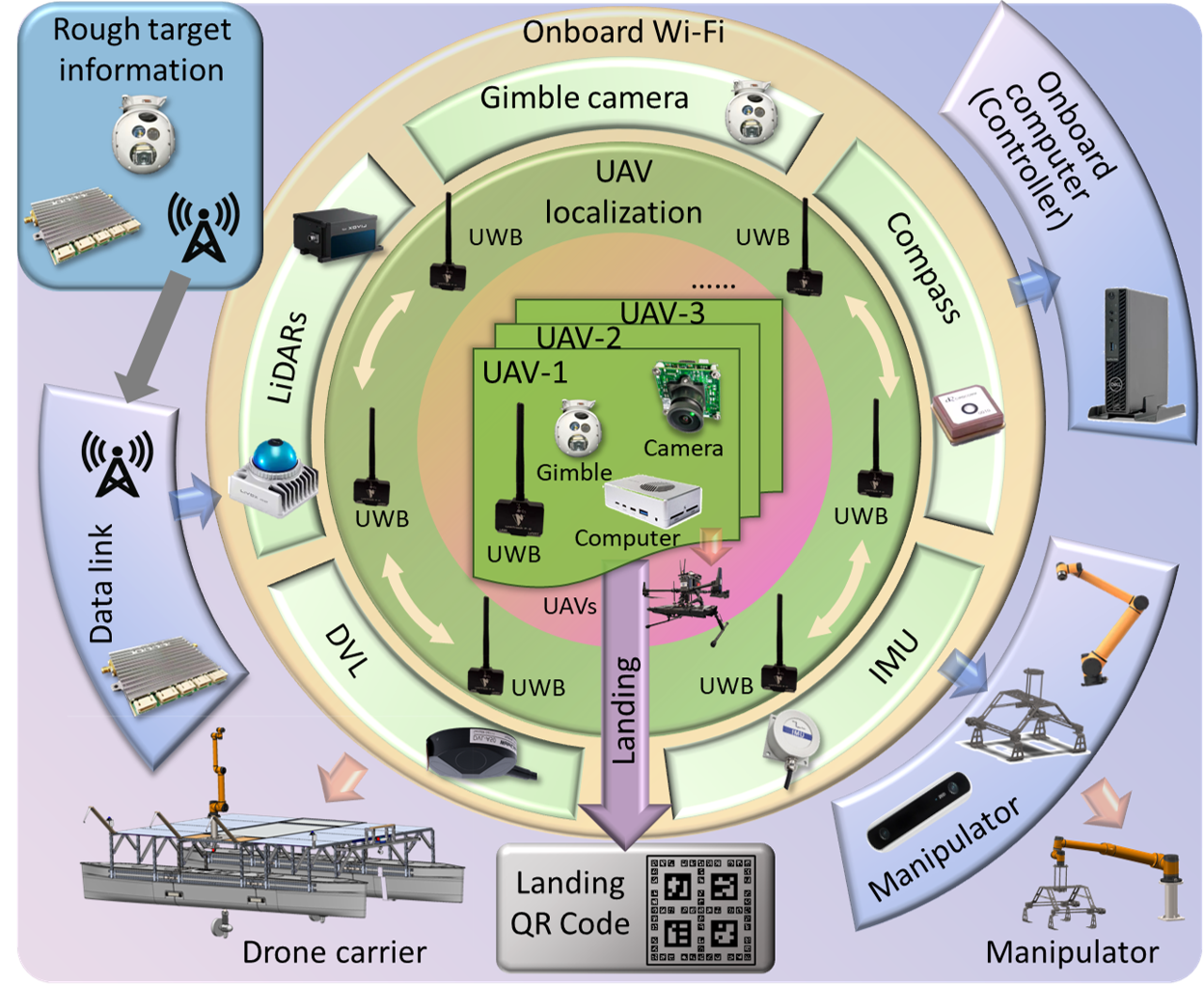} 
		\caption{Composition and relationships of electronic devices on drone carriers}
		\label{fig:sub2}
	\end{subfigure}
	\caption{Modular design of both the (a): catamaran and (b): the drones, manipulator and sensors on the drone carrier.}
	\label{fig:main}
\end{figure*}

\subsubsection{Catamarans component}
The USV system consists of a modular catamaran structure. Due to the inherent stability of catamarans, it can maintain stable navigation even under moderate adverse ocean conditions. The catamaran hull is composed of propulsion modules and non-propulsion modules:
Device Compartment: Includes power batteries, propellers, and steering mechanisms to provide forward propulsion for the USV and enable vector control.
Support Compartment: Composed of hollow chambers with moderate shaping capabilities. The structure is symmetric about its central axis to enhance structural interchangeability and mold reusability.
These modules can be flexibly assembled according to mission requirements, adapting to a variety of tasks.

The two hulls of the catamaran are connected by a truss structure, which can also be adjusted to modify the width of the USV based on the number of drones it carries. The docking side of the USV is equipped with a common docking and securing device. This device uses a spring-adhesion mechanism installed on the underside of the hull to fix the USV to the target vessel, and a mechanical docking hook mounted on the deck to secure the vessel mechanically. A collision detection system is installed at the base of the docking hook, i.e., Fig. \ref{fig:System} (e), and linked to a release pin mechanism. Upon collision, the hook is released and attaches to the edge of the target vessel’s deck. Through research and analysis, the combined docking hook and adhesion mechanism design is suitable for docking with most medium-sized manned and unmanned vessels, enabling autonomous docking in ocean environments.
The modular assembly of the USV can be completed within a day, according to the hull size and mission characteristics. 

\begin{figure*}
	\centering
	\includegraphics[width=1\linewidth]{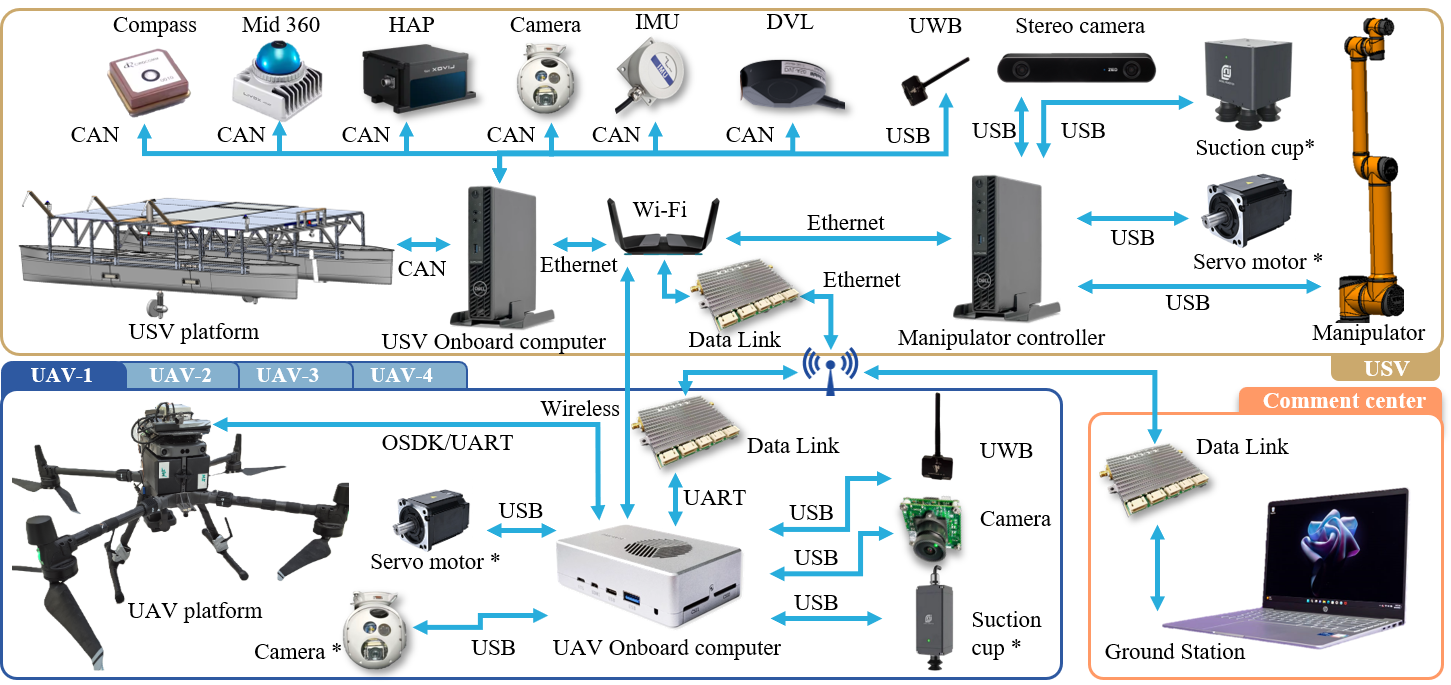}
	\caption{Physical diagram of components in a drone carrier. The subscript ($^*$) represents the alternative payloads for different tasks.}
	\label{fig:hardware}
\end{figure*}

\subsubsection{USV onboard sensors}
All the sensors in a drone carrier is illustrated in Fig. \ref{fig:hardware}. The USV is equipped with various sensors for detection: LiDAR sensors mounted around the hull for 360-degree radar coverage; gimbal for directional coverage and target search; DVL for underwater detection and speed feedback; Compass to output the USV's heading; IMU for attitude measurement; Data links for long-range communication and networking with other drone carriers, UAVs and offshore command centre.

The LiDAR models used in this study are \textit{LIVOX HAP} and \textit{LIVOX MID360}, with point cloud data read via the manufacturer-provided ROS driver. Upon connection to the LiDAR, the driver publishes point cloud data in the \textit{pointcloud2} format in ROS. The point cloud is then processed using the Point Cloud Library (PCL) for operations such as cropping, filtering, transformation, and feature extraction. 
The DVL model utilized is \textit{DVL-A125}, with data acquisition facilitated by the manufacturer-provided ROS driver. 
The onboard computer handles tasks such as reading and processing LiDAR point cloud data, receiving fused state information from IMU and DVL, and controlling the propeller speed and thrust direction of the USV. 

All these onboard devices are networked via a router installed in the hull, forming an onboard network. An onboard computer processes data, records mission parameters, and assigns tasks, relaying critical information back to the command centre.

\subsection{UAVs}

The GNSS-denied search and transport drone for marine environments is based on a mature drone platform, with environmental adaptability improvements incorporated into its design.
To begin with, the drone's landing gear has been widened to accommodate the swaying motion of marine platforms. Anti-slip foot pads have been added to ensure stability on wet and oscillating platforms. Besides, by installing UWB antennas at various positions on the drone's body, the impact of body structure occlusion on UWB antenna signals is effectively reduced, thereby improving positioning accuracy.
Thirdly, landing in conditions involving strong winds, motion, and oscillation introduces significant uncertainty. To address this, the drone integrates anti-interference control logic and a wide-angle landing camera installed at the tail, enabling reliable detection and recognition of landing markers on the deck of the USV at various altitudes and positions. This enhances the success rate of landing and recovery operations.
Last but not least, the multi-functional robotic gripper, designed specifically for the drone carrier system, must operate under the condition of a drone landing accuracy of 20 cm. It can grasp standard containers with dimensions of 30 cm $\times$ 20 cm and a thickness of no more than 10 cm, also maintains a stable grasping performance in high-humidity and high-salinity environments.

\begin{figure*} [h]
	\centering
	\includegraphics[width=6.5in]{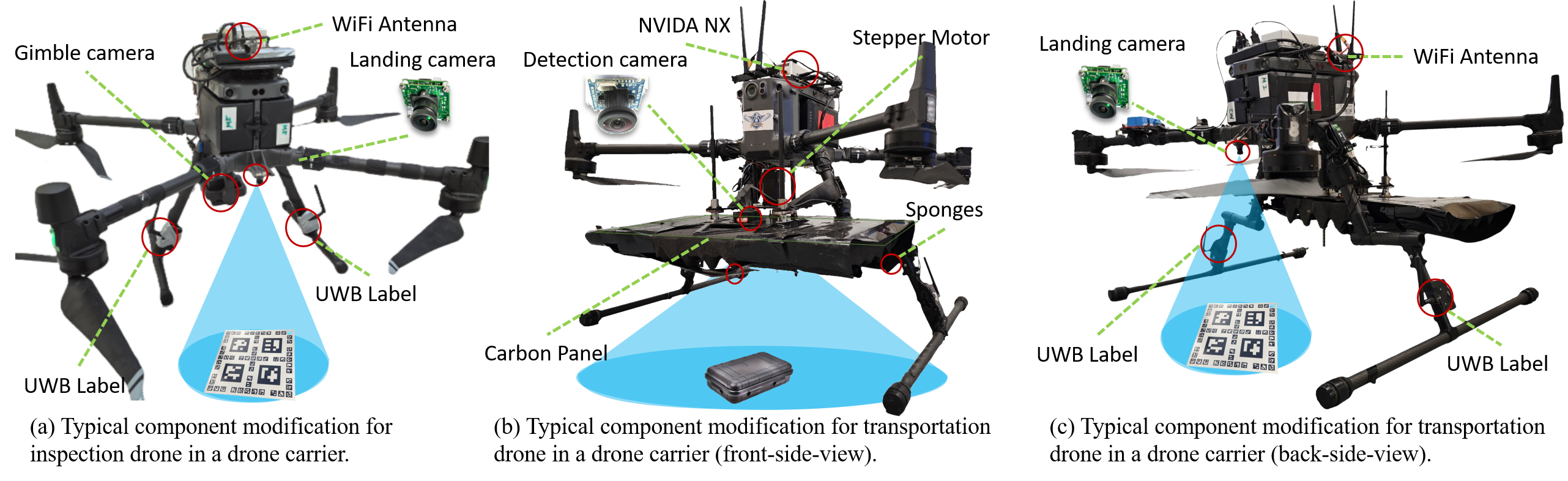}
	\caption{Modification of a typical drone DJI-M300, including localization (UWBs), payloads, landing component (landing camera) and intelligent module (NVIDA NX).}
	\label{fig:UAV}
\end{figure*}

\subsection{Manipulator}
\begin{figure*} [h]
	\centering
	\includegraphics[width=6
	in]{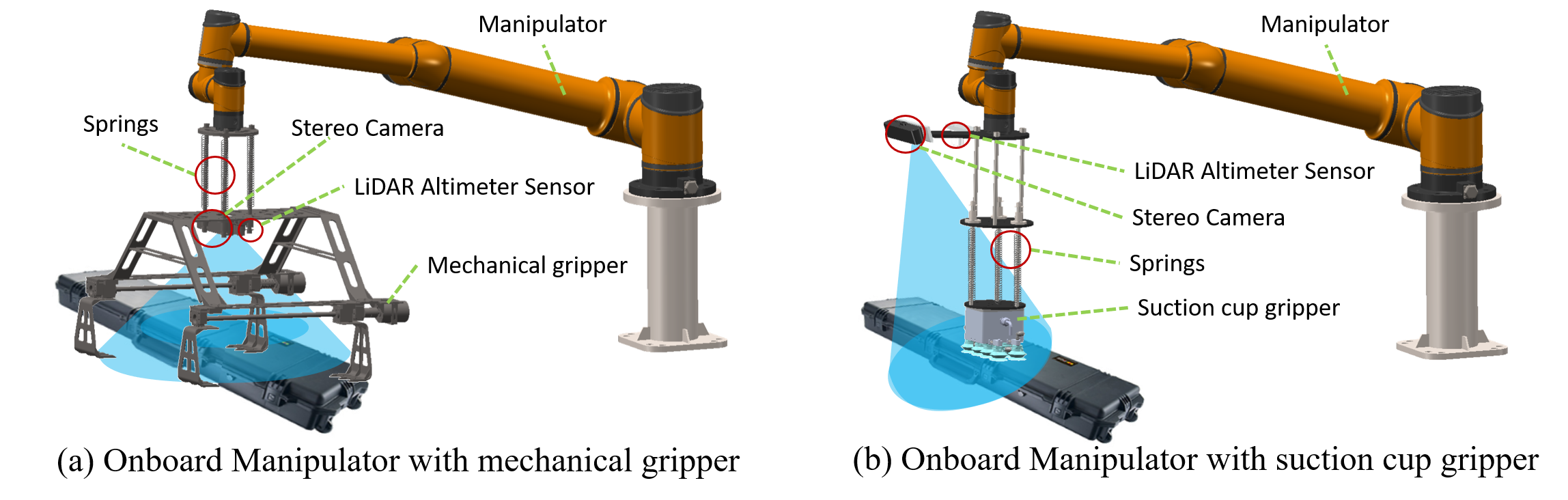}
	\caption{Two typical grippers mounted on the manipulator applied in the marine transportation task. }
	\label{fig:Manipulator}
\end{figure*}

The drone carrier equipped with a robotic arm enables precise target searching within a specific range through a camera mounted on the end of the robotic arm. The arm is also equipped with a gripper to grasp and transport heavier objects. Fig. \ref{fig:Manipulator} illustrated two typical grippers mounted on the manipulator applied in the marine transportation task: the mechanical gripper with stereo motor automatically grasps and clamps the object through the lead screw nut mechanism. The suction cup gripper (EVS08) can grasp large-volume but surface-flat objects, like board or box. However, utilizing a robotic arm for object transportation in marine environments presents several challenges, such as the relative motion and oscillation between vessels, which result in random target movements. These dynamics increase the requirements for the manipulator's operational range and precision.

To ensure the stability of path planning to the greatest extent, the robotic arm's path planning is implemented using the MoveIt package in ROS. Considering that the AUBO robotic arm lacks an API for circular motion, trajectory planning is achieved by sampling points along the trajectory and fitting them with linear segments. Path planning is required during both the target grasping phase and the target returning phase. The MoveIt package is generated based on the URDF model of the robotic arm, abstracting the robot into configuration space (C-Space). The target's pose information, provided by the vision system, is used to invoke the Motion Planning Library (OMPL) to generate motion trajectories for the robotic arm automatically. MoveIt further processes the trajectory points returned by the planner according to the robot's control parameters (e.g., speed and acceleration limits) to produce a complete trajectory, including timestamps, position, velocity, and acceleration information. The trajectory is executed by invoking the AUBO robotic arm's API, reproducing the planned trajectory. 

To ensure stable motion, the robotic arm is controlled in the ROS environment. The URDF file of the robotic arm is used to generate the MoveIt package, and the control program relies on multiple function packages to interact with the AUBO robotic arm's API. Custom coordinate transformation matrix functions are developed to achieve precise control of the robotic arm's trajectory and orientation. Self-collision avoidance is implemented by modifying the URDF file to include data for the end effector, camera base, and camera. A comprehensive URDF file is generated, enabling extensive sampling of the robotic arm's motion using MoveIt to avoid self-collisions. Environmental collision avoidance also leverages the MoveIt package. MoveIt provides a planning scene monitoring module that detects obstacles within the robot's environment. By modeling and exporting environmental obstacles, collision-free paths can be planned. Target grasping relies on the EVS08 suction cup or a custom-developed grasping mechanism. The robotic arm controls the suction cup's position and orientation relative to the target object and invokes specific functions to perform grasping. Through precise end-effector pose, velocity, and acceleration control, the stability of the grasping process is ensured. The vision system provides the "grasping position relative to the camera coordinate system." The robotic arm is controlled to move to the corresponding pose, with the target's information indicating a vertically upward pose. The robotic arm's control algorithm executes a top-down tracking motion, driving the suction cup to complete the grasping task.

The installation schematic of the robotic arm on the USV is shown in Fig. \ref{fig:System} (c). To balance the weight, the power supply and control cabinet for the robotic arm is installed on the opposite side of the USV. The robotic arm is equipped with a stereo camera, a spring-loaded gripper mechanism, vacuum suction cups, and waterproof adhesive, enabling the identification and manipulation of target objects in complex marine environments.
\begin{figure} [h]
	\centering
	\includegraphics[width=0.5\textwidth]{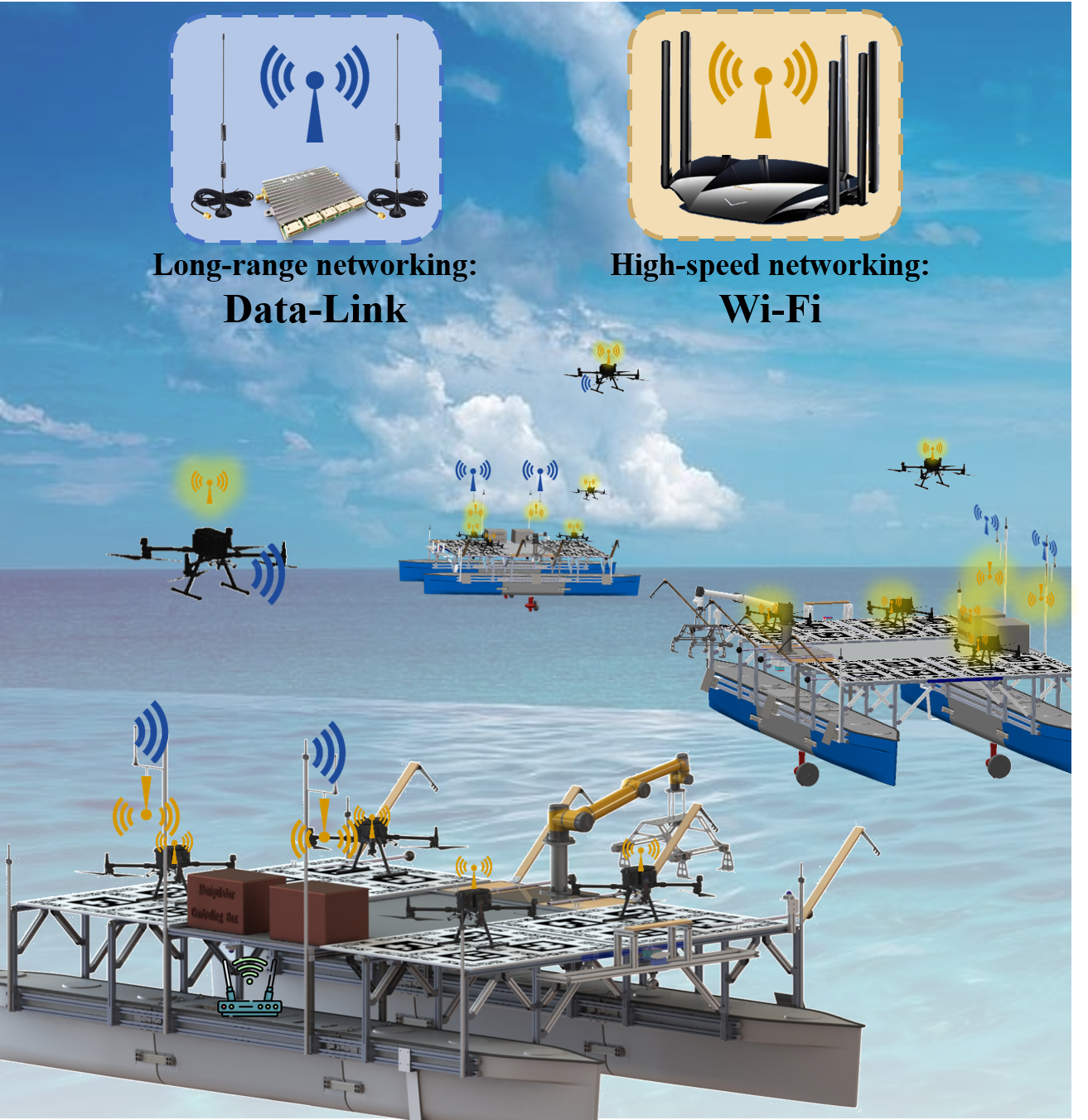}
	\caption{Multiple robotics communication using data-link for long-range networking and Wi-Fi for close high-speed communication inside the drone carriers. }
	\label{fig:communication}
\end{figure}

\subsection{System Communication}
The USV communicates with nearby drones through onboard Wi-Fi and with distant drones via a data link. The manipulator's control cabinet is connected to the onboard router via an Ethernet cable, integrating the robotic arm into the system for sending and receiving commands. UWB modules of varying heights are installed at the corners of the deck, fused with vessel pose information to establish a coordinate system with the USV’s centre as the origin and the forward-left-up directions as the axes. By measuring the position of the UWB module on the drone within this coordinate system, the drone's local positioning is achieved.

In addition, each drone landing position is equipped with a QR code composed of differently-sized patterns. These QR codes provide precise landing position information for drones of varying altitudes, enabling stable landings despite platform fluctuations and wave disturbances.

\section{Autonomous Approaching, Docking and Transportation in GNSS-denied Sea Exnvironment}

\begin{figure*} [t]
	\centering
	\includegraphics[width=6in]{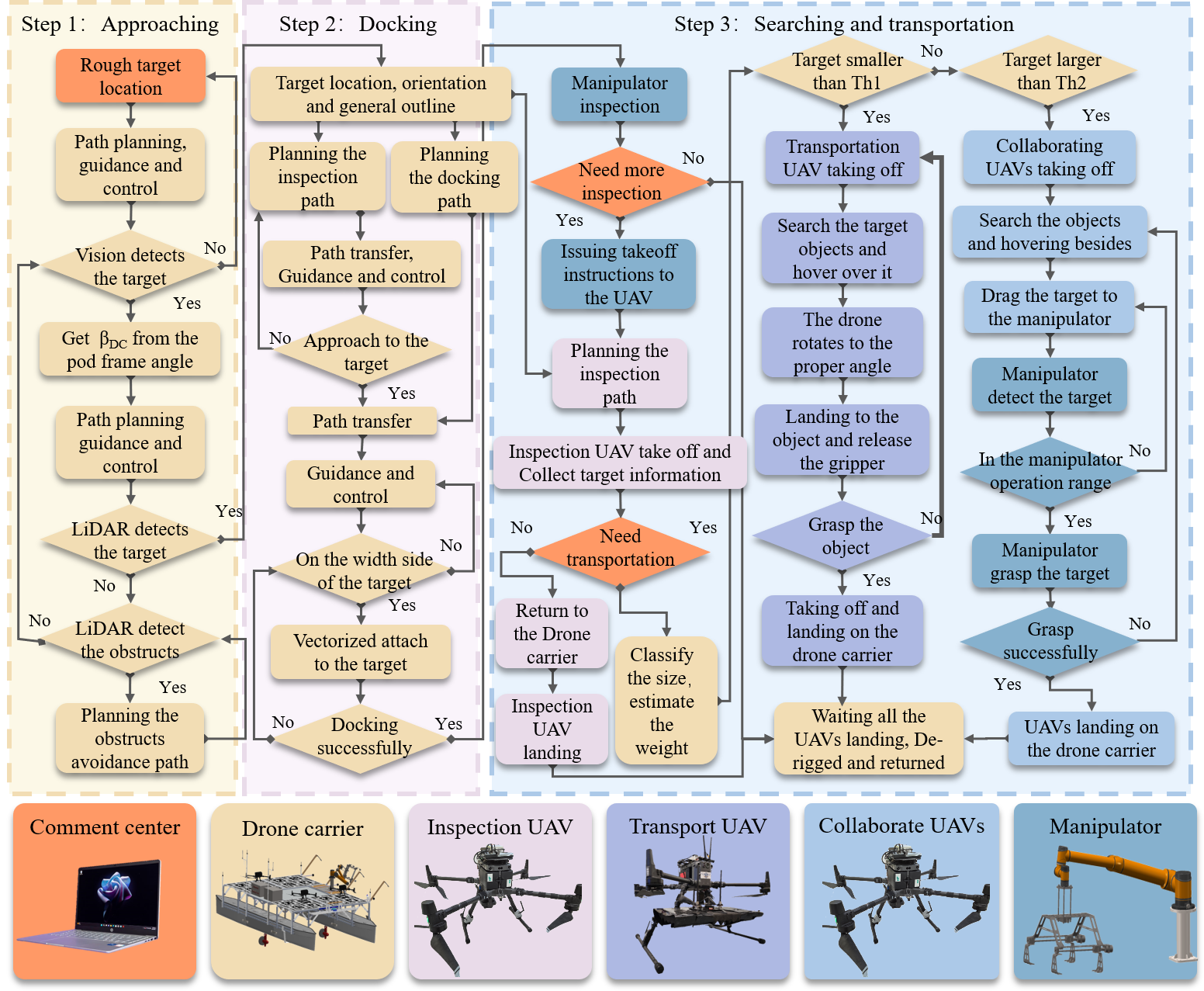}
	\caption{Flow chat for the drone carrier conducting wide-range inspection and transportation in GNSS-denied marine environment.}
	\label{fig:Software}
\end{figure*}

The operational flowchart for the drone carrier performing wide-range inspection and transportation tasks is presented in Fig. \ref{fig:Software}. Before we expand the discussion of the methodology, a few frames used in the whole process need to be defined. The East-North-Up (ENU) frame is the inertial system, where 
$x$, $y$ and $z$ respectively represent the East, North and Up. Front-Left-Up (FLU) frame is according to the robotics system and defined by the front, lift and upward axis of the robotic's body. The notation of each label is illustrated in Tabal \ref{tab:frm}.

\begin{table}[h]
	\centering
	\caption{Coordinate frames define.}
	\begin{tabular}{ccc}
		\hline
		Notation & Frame  & Description  \\\hline
		$I$   & ENU &   The ground        \\
		$TV$  & ENU &   Target vessel \\
		$GC$  & FLU &   Gimbal camera     \\
		$DC$  & FLU &   Drone carrier     \\
		$D_i$  & FLU &   The $i^{th}$ UAV  \\
		$O_j$  & FLU &   The $j^{th}$ object\\
		$MA$  & FLU &   Manipulator       \\
		\hline
		
	\end{tabular}\label{tab:frm}
\end{table}

\subsection{Dynamic Model of the Drone Carrier}

As illustrated in Fig. \ref{fig:main} (a), the drone carrier has two thrusters mounted on the middle of both device compartment hulls. the dynamic model governs the vessel's movement based on hydrodynamic forces, inertial properties, and external actuation, which can be expressed as:

\begin{equation}
	\mathbf{M}\dot{\mathbf{v}}_{DC}+\mathbf{D}\textbf{v}_{DC}=\boldsymbol{\tau}_{act}
	\label{eq:1}
\end{equation}
where \textbf{M} and \textbf{D} represent the Inertia and Damping matrix, respectively. $\textbf{v}_{DC}$ is the velocity vector of the drone carrier, including the surge, sway, and yaw rate. The $\textbf{D}$ is the first-order hydrodynamic coefficient matrix:

\begin{equation}
	\mathbf{D}=\left[\begin{array}{ccc}
		D_{x} & 0 & 0 \\
		0 & D_{y} & 0 \\
		0 & 0 & D_{z}
	\end{array}\right]
	\label{eq:3}
\end{equation}
where subscripts \textit{x}, \textit{y}, \textit{z}, respectively, represent the direction in surge, sway and yaw, under the USV FLU frames. The actuation forces $\mathbf{\tau}_{act}$ in Eq. \eqref{eq:1} are generated by the thrusters. In a modularized drone carrier with two thrusters, these forces and moments can be written as:
\begin{equation}
	\boldsymbol{\tau}_{\mathrm{act}}=\left[\begin{array}{c}
		F_{x} \\
		F_{y} \\
		M_{z}
	\end{array}\right]=\mathbf{B}\left[\begin{array}{l}
		T_{1} \\
		T_{2}
	\end{array}\right]
	\label{eq:4}
\end{equation}
and
\begin{equation}
	\mathbf{B}=\left[\begin{array}{cc}
		\cos \left(\theta_{1}\right) & \cos \left(\theta_{2}\right) \\
		\sin \left(\theta_{1}\right) & \sin \left(\theta_{2}\right) \\
		d_{1}\cos \left(\theta_{1}\right) & -d_{2}\cos \left(\theta_{2}\right)
	\end{array}\right]\left[\begin{array}{l}
		T_{1} \\
		T_{2}
	\end{array}\right]
	\label{eq:5}
\end{equation}
where \textbf{B} is the is the control input matrix that maps the thrust forces to the surge, sway, and yaw forces. $T_1$ and $T_2$ are the forces generated by the two thrusters, while $\theta_1$ and $\theta_2$ are the angles of the two thrusters relative to the centre line of the USV. Since the two thrusts are amounted symmetrically on the USV, $d_{1}=d_{2}=d$ represents the distances of each thruster from the centre of mass of the USV. For the Simplified Control Model, where the two thruster only generates a force in the surge direction, $\theta_1=\theta_2=0$, the actuation force can be expressed as:
\begin{equation}
	\boldsymbol{\tau}_{\mathrm{act-s}}=\left[\begin{array}{c}
		T_{1}+T_{2} \\
		0 \\
		dT_{1}-dT_{2}
	\end{array}\right]
	\label{eq:6}
\end{equation}

The simplified control model cannot provide the sway force that can only be used in the long-range navigation and approach. For the attaching or the docking process, we use the Thruster Rotation Model. The actuation force can be expressed as:
\begin{equation}
	\boldsymbol{\tau}_{\mathrm{act-r}}=\left[\begin{array}{c}
		T_{1}\cos \left(\theta_{1}\right)+T_{2}\cos \left(\theta_{2}\right) \\
		T_{1}\sin \left(\theta_{1}\right)+T_{2} \sin \left(\theta_{2}\right) \\
		dT_{1}\cos \left(\theta_{1}\right) -dT_{2}\cos \left(\theta_{2}\right)
	\end{array}\right]
	\label{eq:6}
\end{equation}

\subsection{Step 1: Offshore Navigation and Approaching}\label{chpB}

The offshore localization and navigation in a GNSS-denied environment are achieved using an onshore gimbal camera, as illustrated in Fig. \ref{fig:main}(d). While the detailed methodology is discussed in a previous study \cite{akram2024long}, this work integrates the entire localization process, linking the onshore camera to the target vessel. The approaching process can be divided into two phases. 

\begin{figure} [h]
	\centering
	\begin{subfigure}[b]{0.48\textwidth}
		\centering
		\includegraphics[width=\textwidth]{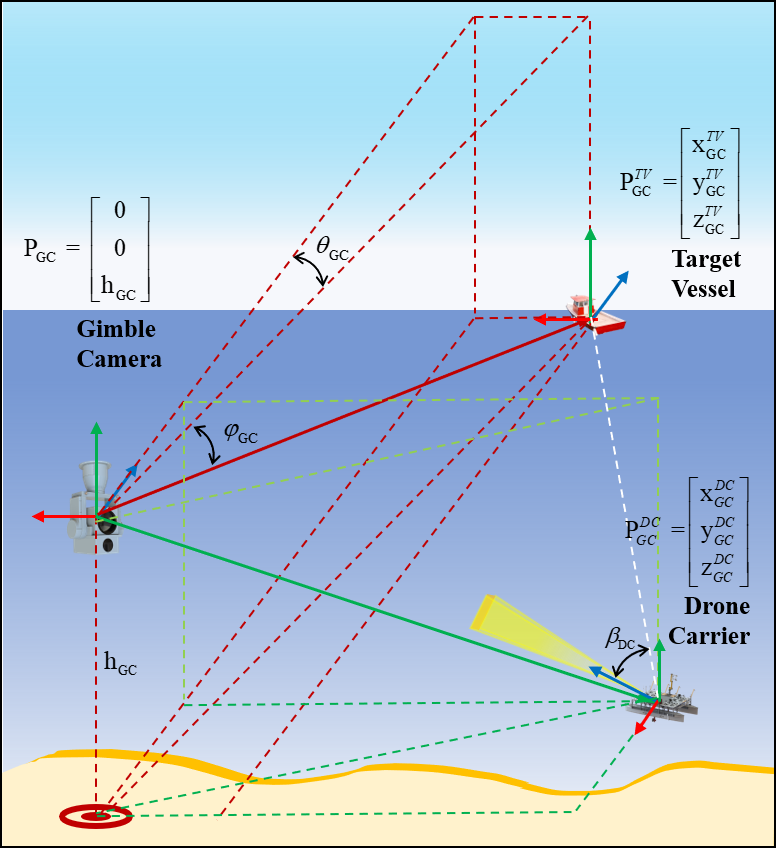}
		\caption{Target position in the onshore gimbal camera frame.}
		\label{fig:tri1}
	\end{subfigure}
	\begin{subfigure}[b]{0.48\textwidth}
		\centering
		\includegraphics[width=\textwidth]{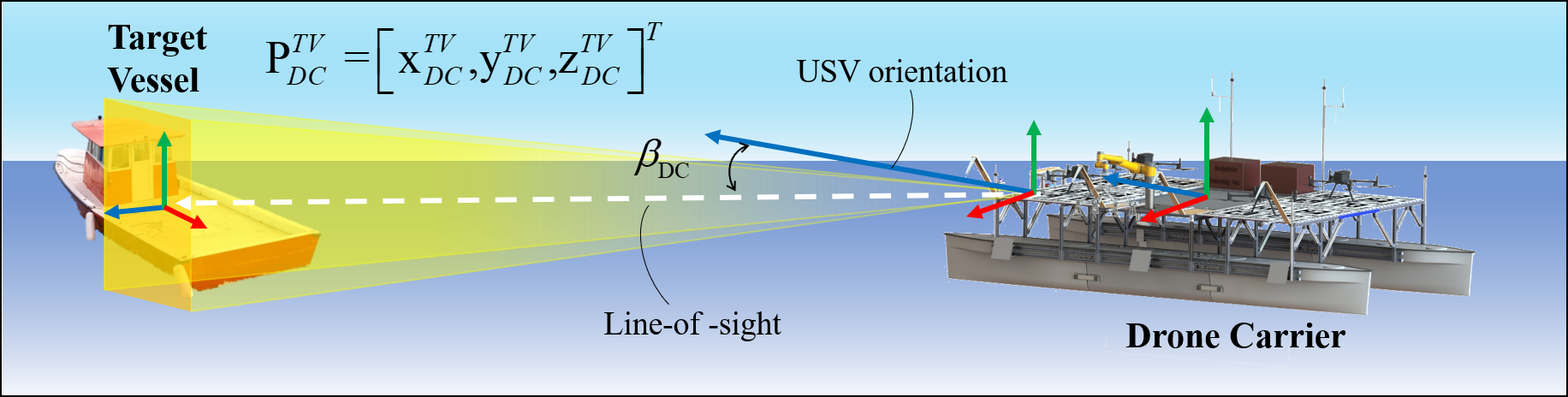}
		\caption{Target position in drone carrier frame.}
		\label{fig:another}
	\end{subfigure}
	\caption{Illustration of the target localization using: (a) onshore gimbal camera, and (b) USV onboard gimbal camera.}
	\label{fig:tri2}
\end{figure}

\textbf{Onshore gimbal camera navigation:} Initially, an onshore 2-axis gimbal camera detects both the drone carrier and the target. This shore-based gimbal camera can be installed on tall towers or drones hovering along the shoreline. The drones use downward-facing vision to autonomously localize the camera, ensuring precise determination of the camera's position. Using YOLOv5 \cite{yolov5}, the system autonomously identifies target vessels and USV. By adjusting the gimbal Euler angles, the identified targets are kept centered within the frame. The target's position is then calculated using the trigonometric relationships depicted in Fig. \ref{fig:tri2}(a).

The position of the gimbal Camera can be predefined as $\textbf{P}_\text{GC} = [0,0,h_\text{GC}]^T$ in the FLU frame. Given the camera's Euler angles in the \textit{GC} frame, denoted by $\theta_\text{GC}$ and $\varphi_\text{GC}$, indicate the horizontal and vertical angles of the camera, the target position under GC frame, ${\textbf{P}}_{GC}^{{\rm{TV}}}$, can be computed by:
\begin{equation}
	{\textbf{P}}_{GC}^{{\rm{TV}}}{\rm{  = }}\left[ {\begin{array}{*{20}{c}}
			{{\rm{x}}_{GC}^{{\rm{TV}}}}\\
			{{\rm{y}}_{GC}^{{\rm{TV}}}}\\
			{{\rm{z}}_{GC}^{{\rm{TV}}}}
	\end{array}} \right] = h\left[ {\frac{{\sin \left( {{\theta _{GC}}} \right)}}{{\tan \left( {{\varphi _{GC}}} \right)}},\frac{{\cos \left( {{\theta _{GC}}} \right)}}{{\tan \left( {{\varphi _{GC}}} \right)}},0} \right]^T
	\label{eq:7}
\end{equation}

It is important to note that when state variables are derived solely from geometric relationships, the measurements between frames are uncorrelated, leading to considerable variability and uncertainty. To mitigate this, we implement the Extended Kalman Filter (EKF), which effectively integrates information across multiple frames while accounting for nonlinear dynamics, thereby enhancing estimation accuracy and robustness. Using the estimated USV orientation, $\beta_{DC}$, the USV navigates towards the target vessel by calculating the yaw angle between the USV and the target. 
\begin{equation}
	{\beta _{DC}} = \theta _{DC}^{IMU}-act\left( {\frac{{{\rm{y}}_{GC}^{{\rm{TV}}} - {\rm{y}}_{GC}^{{\rm{DC}}}}}{{{\rm{x}}_{GC}^{{\rm{TV}}} - {\rm{x}}_{GC}^{{\rm{DC}}}}}} \right) 
	\label{eq:8}
\end{equation}
where $\theta _{DC}^{IMU}$ is the yaw detected by USV onboard IMU. This navigation process achieves an accuracy of up to 200 meters within a 3 km offshore range.

\textbf{USV onboard gimbal camera navigation:} In the second approach phase, the onboard gimbal camera and LiDAR utilize the prior images of the target to identify and compare maritime objects, as illustrated in Fig. \ref{fig:tri2}(b). Non-target objects encountered along the planned path are classified as obstacles, and the system adjusts the route to avoid them.
The USV onboard gimbal camera can identify and compare targets at a maximum distance of 500 meters. Once the camera locks onto the target, the USV’s navigation transitions to a vision-guided phase driven by the USV onboard gimbal camera: ${\beta _{DC}} = \theta _{DC}^{pod}$. The pod’s frame angle is used as the heading angle for the USV, enabling it to navigate toward the target. 

\begin{figure*} [htb!]
	\centering
	\begin{subfigure}[b]{0.46\textwidth}
		\includegraphics[width=\textwidth]{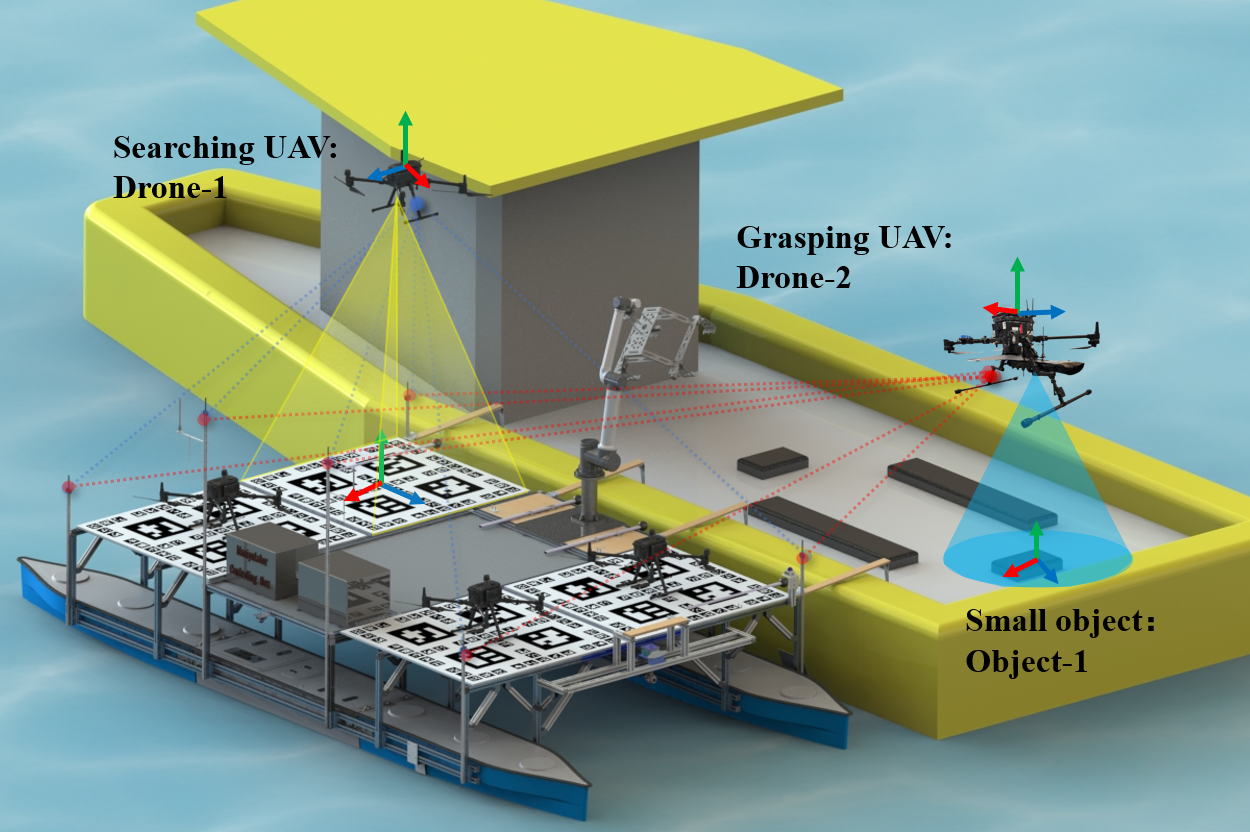}
		\caption{Small object localization and grasping.}
		\label{fig:smallobj}
	\end{subfigure}
	\begin{subfigure}[b]{0.46\textwidth}
		\includegraphics[width=\textwidth]{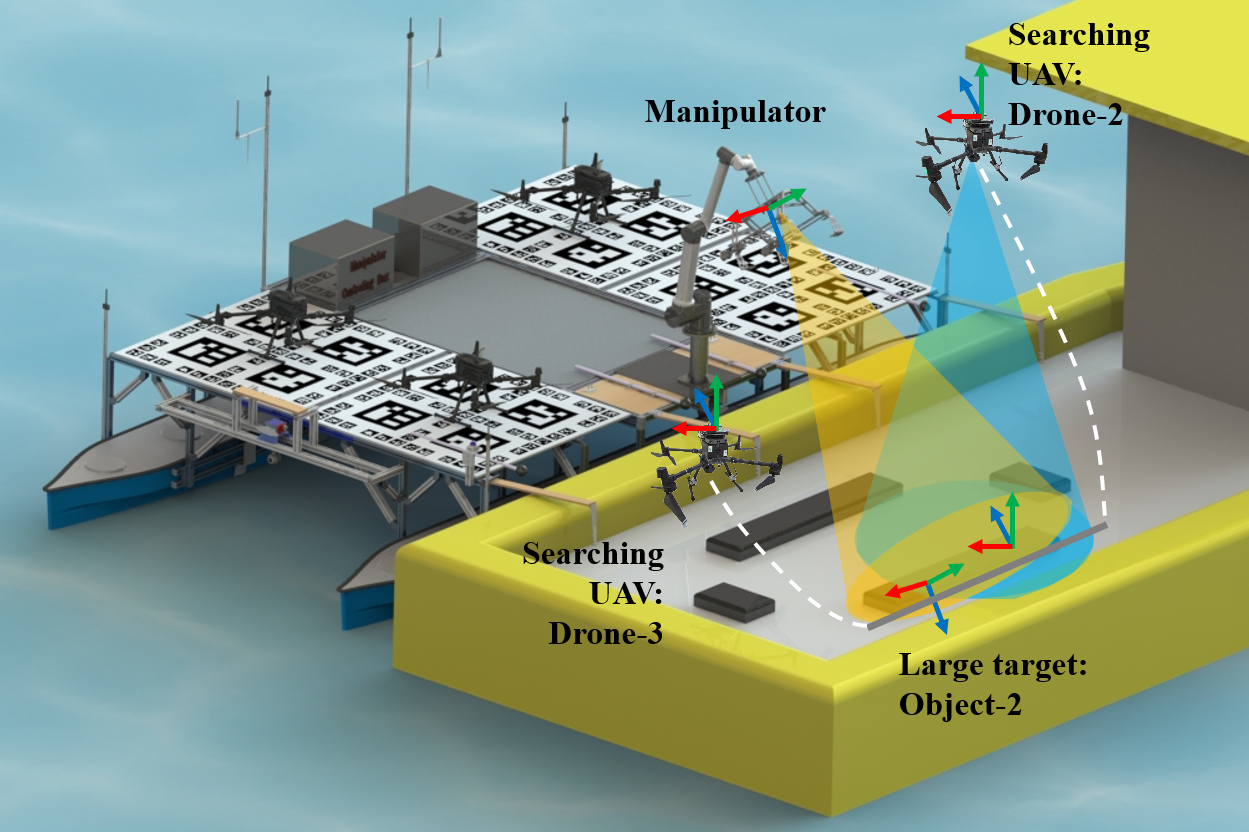}
		\caption{Collaborate big object localization and grasping.}
		\label{fig:bigobj}
	\end{subfigure}
	\caption{Collaborate object localization and grasping using drones and a manipulator, where the drones are localized based on QR code and UWB in a GNSS-Denied sea environment.}
	\label{fig:obj}
\end{figure*}

\begin{figure*} [htb!]
	\centering
	\includegraphics[width=\textwidth]{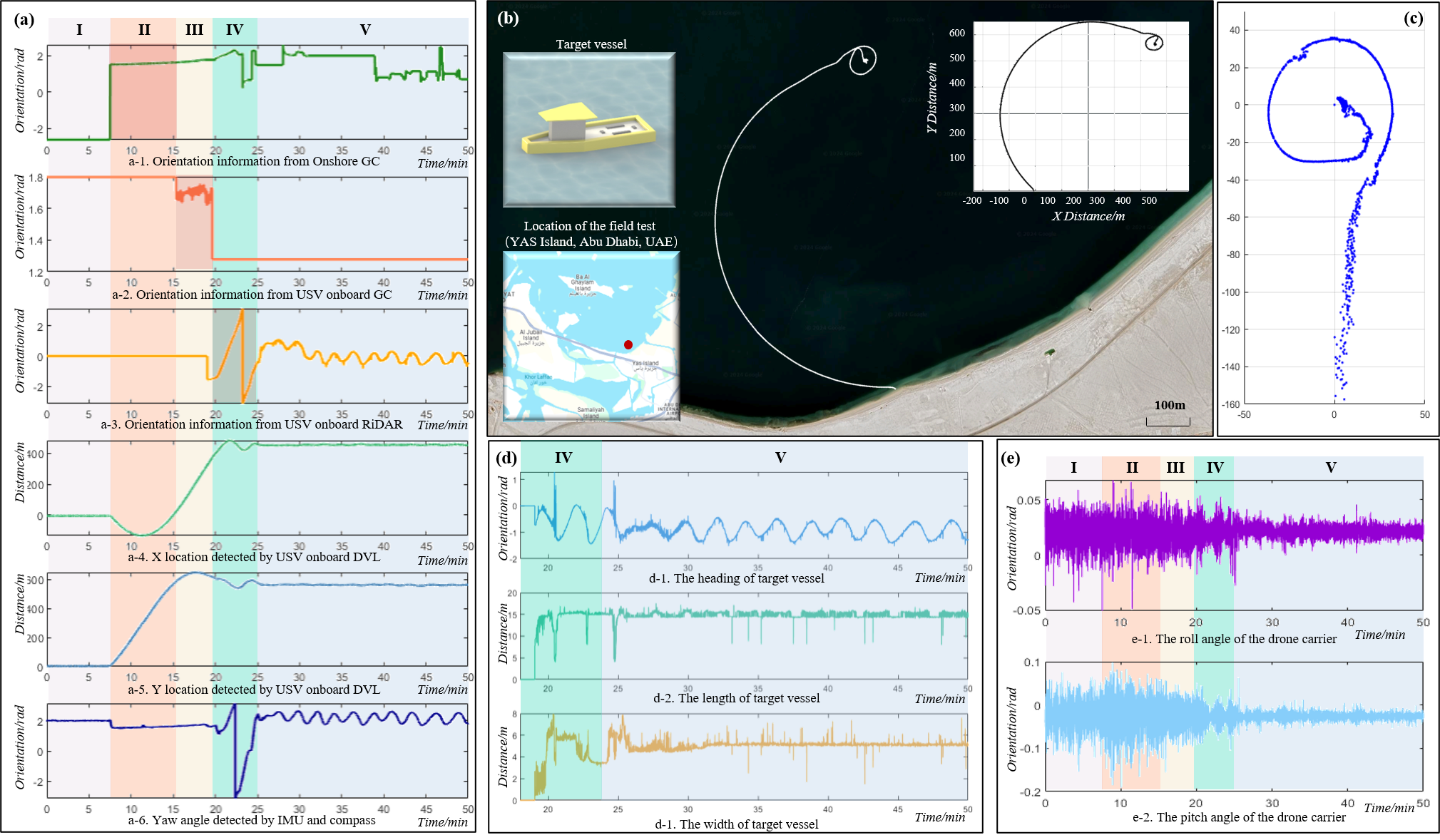}
	\caption{Field test data of the drone carrier in a GNSS-denied sea environment: (a) Desired heading angle and position of the drone carrier during the approaching phase. (b) Routine path measured by the DVL in the field test area. (c) Target position in the \textit{DC} frame, as measured by LiDAR. (d) Onboard LiDAR measurements of the target vessel's orientation, length, and width. (e) Roll and pitch angles of the drone carrier throughout the test process.}
	\label{fig:fieldtest}
\end{figure*}

\begin{figure*} [htb!]
	\includegraphics[width=\textwidth]{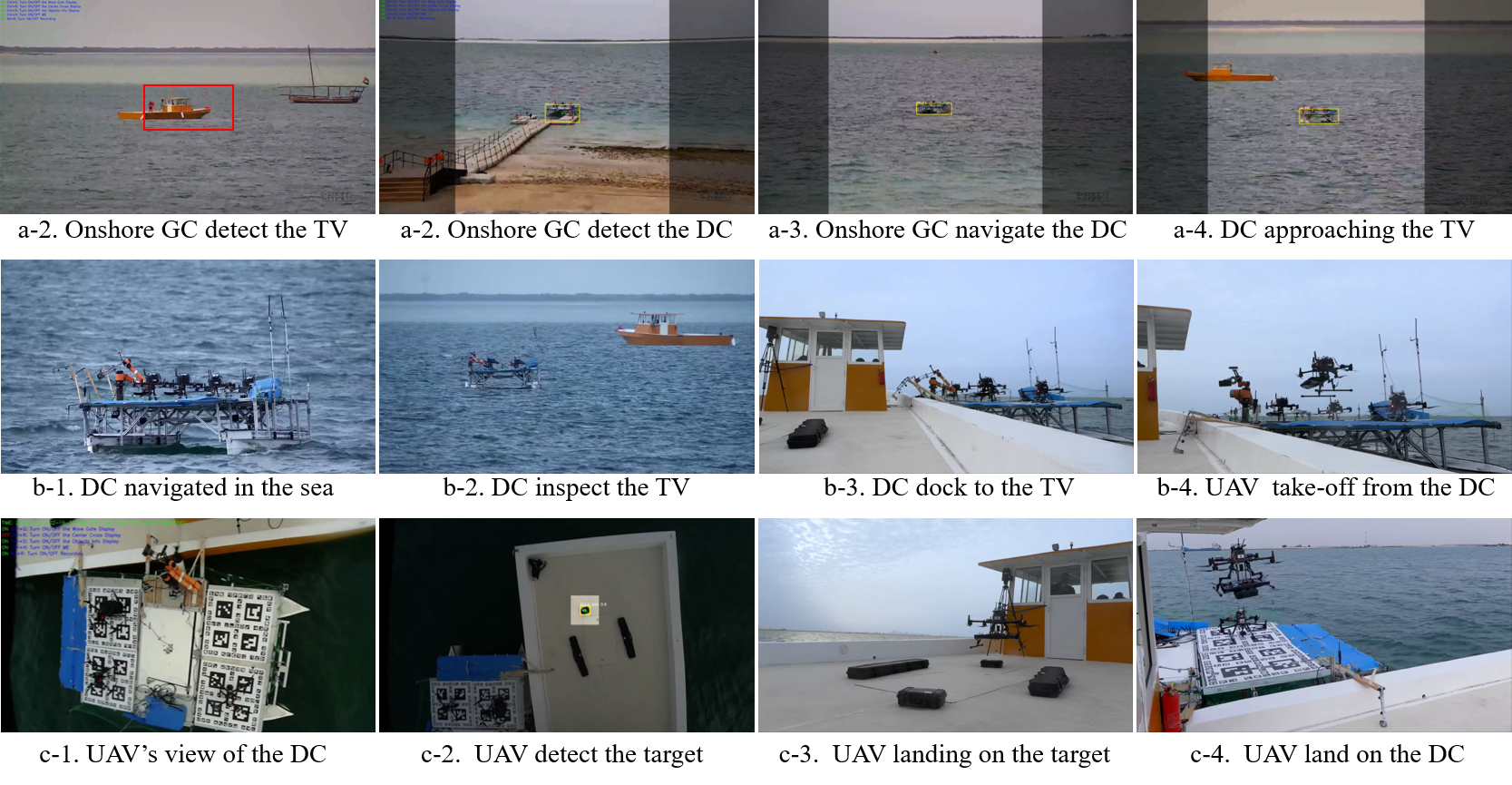}
	\caption{Field test images of the drone carrier in a GNSS-denied sea environment: (a) Target vessel localization and drone carrier navigation based on an onshore gimbal camera. (b) Autonomous approaching and docking to a target vessel. (c) UAV takes off, search and grasps the object on the target vessel, and lands back to the drone carrier.}
	\label{fig:realtestimag}
\end{figure*}

\subsection{Approaching and Docking}
As described in Sec. \ref{chpB} and  \ref{chpC}, the entire approaching and docking process is divided into five phases in the field test:
\begin{itemize}
	\item \textit{I}: Preparation Phase
	\item \textit{II}: Onshore GC Guidance Phase
	\item \textit{III}: USV Onboard GC Guidance Phase
	\item \textit{IV}: Measurement and Docking Phase
	\item \textit{V}: Docking Complete Phase
\end{itemize}

\subsection{Step 2: Autonomous Docking}\label{chpC}

The onboard LiDAR of the USV has a maximum detection range of 200 meters. Upon identifying the target vessel using point cloud data, the USV autonomously generates a reconnaissance path, maintaining a circular trajectory around the target at a 50-meter radius. During this process, the USV scans and models the target, producing detailed 3D dimensional and orientation data of the vessel. While this step has been thoroughly described elsewhere, a brief overview is provided here:

\textbf{Measurement:} Five LiDARs are strategically placed to capture complete point clouds without blind spots, employing approximate time synchronization at 5 Hz to ensure accurate measurements. Filtered point clouds remove noise, low-intensity points, and outliers, and are segmented into clusters using Euclidean distance and KD-Tree search methods. Clusters are enclosed by optimal rectangles using a variance-based fitting algorithm that minimizes squared errors. For dock alignment, an L-shape fitting algorithm \cite{Zhang2017} determines the heading angle by modeling the dock as two perpendicular lines, enhancing navigation accuracy in real-world environments. 

\textbf{Navigation:} 
We utilize the Dubins curve \cite{Cai2023} to design an autonomous docking path for the USV \cite{Cai2023}. We segment the curve into stages with variable turning radii to enhance flexibility, constrained by the USV’s weight, length, and speed range. Assuming minimal turning radius \textbf{$\psi$} is proportional to the current velocity $V_i$, and neglecting roll and pitch effects.

\textbf{Docking:} 
After completing the 3D modeling of the target vessel, the system transitions to the docking phase. This phase primarily involves mechanically connecting the USV to the target vessel to release a UAV for reconnaissance and transport tasks. During the circling phase, the USV selects the long side of the target vessel for docking. Using LiDARs to monitor the target vessel's position in real-time, the USV manoeuvres toward the selected side. The USV utilizes vector propulsion for lateral movement to approach the target vessel. When the USV contacts the target vessel, a mechanical docking hook is automatically released to establish a mechanical connection between the two vessels.

Once the LiDAR confirms the USV is securely positioned alongside the target vessel, the system will determine that docking is successful and proceed to the detailed search and transport phase.

\begin{table}[]
	\caption{Drones, manipulate and objects defined in Fig. \ref{fig:obj}}
	\centering
	\begin{tabular}{lll}
		\hline
		Item  & Position & Description \\ \hline
		\begin{tabular}[c]{@{}l@{}}Searching UAV:\\ Drone-1\end{tabular}  &  ${\bf{P}}_{DC}^{{ D}_1}$ & Drone-1 position in DC frame \\
		\begin{tabular}[c]{@{}l@{}}Grasping UAV:\\ Drone-2\end{tabular}   & ${\bf{P}}_{DC}^{{ D}_2}$ & Drone-2 position in DC frame \\
		Manipulator & ${\bf{P}}_{DC}^{ MA}$& Manipulator position in DC frame\\
		\begin{tabular}[c]{@{}l@{}}Small object:\\ Object-1\end{tabular}  & ${\bf{P}}_{{ D}_2}^{{{O}}_1}$ & Object-1 position in Drone-2 frame\\
		\multirow{2}{*}{\begin{tabular}[c]{@{}l@{}}Large object:\\ Object-2\end{tabular}} &${\bf{P}}_{{ D}_1}^{{{O}}_2}$& \multirow{2}{*}{\begin{tabular}[c]{@{}l@{}}Object-2 position in both Drone-1\\ and manipulator frame\end{tabular}}  \\ 
		& ${\bf{P}}_{ MA}^{{{O}}_2}$\\ \hline
	\end{tabular}
	\label{tab:position}
\end{table}

\subsection{Step 3: Searching and Transporting}

The search and transportation process of the drones in UWBs frame is illustrated in our previous works \cite{sun2024aerial,chenheterogeneous}. Here we link the methodology of automatic and localization into a whole process. This process can be divided into two parts: namely localization and grasping. 

\textbf{UAV and target localization}:
As illustrated in Fig. \ref{fig:obj}(a), a robotic arm mounted on the USV uses a stereoscopic camera at its end to scan the target vessel. A rough object position under a manipulator frame ${\bf{P}}_{ MA}^{{{O}}_j}{{  = }}{\left[ {{{x}}_{ MA}^{{{O}}_j}{{,y}}_{ MA}^{{{O}}_j}{{,z}}_{MA}^{{{O}}_j}} \right]^T} $ is transfer into the drone carrier frame:
\begin{equation}
	{\bf{P}}_{DC}^{{{O}}_j}{{  = }}{\bf{R}}_{DC}^{{{MA}}}{\bf{P}}_{ MA}^{{{O}}_j}
	\label{eq:9}
\end{equation}
where ${\bf{R}}_{DC}^{{{MA}}}$ is the rotation matrix from the stereoscopic camera (endpoint of manipulator) to the drone carrier frame. The location is sent to the searching drone and takes off using the QR code according to the works from \cite{ref:TAD2024}.
Once the QR code is out of the camera's view, the UAV switches to UWB-based localization, calculating its position via trilateration \cite{ref:FRI2004}. To enhance stability, an Extended Kalman Filter (EKF) fuses IMU data for smoother localization \cite{ref:DKI2020}, while singular elimination and mean value filtering refine target positions \cite{ref:POF1993}. Velocity estimation, $\dot{c}_b$, employs a Kalman Filter \cite{ref:FAC1995}. Addressing the Coverage Path Planning (CPP) problem for the deck, a spiral pattern is adopted for efficient coverage \cite{ref:TSD2019}. UAV movement follows rigid body dynamics \cite{ref:TGI2010}, with a PID controller mitigating nonlinear disturbances and marine perturbations \cite{ref:FUM2022}.

\textbf{Object grasping}:
The grasping UAV autonomously descends above the target object, deploying a mechanical gripper to establish a secure connection with the object. Once secured, the UAV lifts off and uses UWB for real-time positioning. The UAV identifies a QR code on the USV to facilitate precise landing.

For objects that cannot be transported by a single UAV, coordinated transport involving multiple UAVs and a robotic arm is employed. As illustrated in Fig. \ref{fig:obj}, the manipulator allocates the big object and obtains the position of the big object under the MA frame: ${\bf{P}}_{MA}^{{{O_2}}}$. This position is converted to the searching drone frame by a rotation matrix by Eq. \eqref{eq:9}. Two searching UAVs are taking off and navigating by the localization information from the fusion of UWBs and USV onboard IMU. Using the vision-based object caption, the big object's position is allocated under the frame of each UAV, ${\bf{P}}_{D_2}^{{{O_2}}}$ and ${\bf{P}}_{D_3}^{{{O_2}}}$. Each UAV hovers at one end of the large object, lowers its altitude, and positions a flexible tether attached to both UAVs along one side of the object. The UAVs then collaboratively drag the large object towards the manipulator. Meanwhile, the manipulator assesses the distance to the object by visual sensors. Once the object is within the arm’s operational range, the robotic arm manipulates it and transports it onto the USV.

\begin{figure*} [htb!]
	\includegraphics[width=\textwidth]{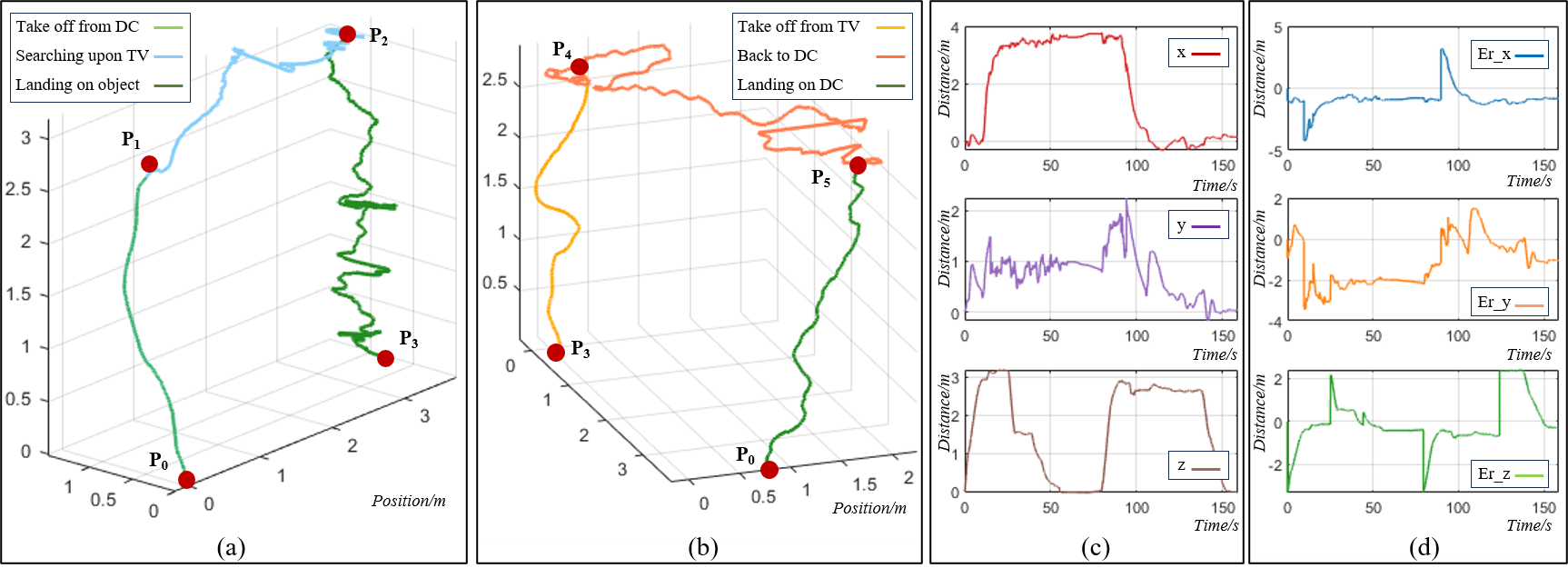}
	\caption{Path and positional trajectory of the UAV transportation object from a target vessel to the drone carrier, where (a) illustrated the 3D position during the taking off from the drone carrier to the object and back to the drone carrier (b), searching and landing on the target. (c) is the 3D position during the transportation process and the position difference between the comment is plotted in (d).}
	\label{fig:dronepath}
\end{figure*}

\section{Field Test and Result}
The entire field test is performed in the waters near Yas Island, Abu Dhabi, under strict supervision by a third party (@ASPIRE). The test area spanned 3 square kilometres and included one target vessel and seven interference vessels (as illustrated in Fig. \ref{fig:fieldtest} (b)). To simulate a GNSS-denied environment, all GPS antennas on the equipment, including the antenna on the DJI-300 drone, were removed. Fig. \ref{fig:fieldtest}(a) illustrates the complete path of the drone carrier during the approach and docking phases.

Fig. \ref{fig:fieldtest} (a) shows the source of the drone carrier's heading information and its corresponding position and heading during these phases. In the subsequent three phases, the expected heading of the drone carrier is provided by the onshore gimbal camera (Fig. \ref{fig:fieldtest} (a-1)), the USV onboard gimbal camera (Fig. \ref{fig:fieldtest} (a-2)), and radar (Fig. \ref{fig:fieldtest} (a-3)). The drone carrier’s onboard control algorithm adjusts the speed of the two motors while continuously correcting its heading to approach the desired heading, ultimately achieving the control objective. In phase \textit{V}, the drone carrier attaches to the target vessel with the docking hook, but still rolling under the sea wave.
The data indicate the reality of the test result and the proposed navigation framework for the drone carrier can successfully operate in a GNSS-denied environment.

Once the onboard radar captures the target vessel, the radar displays the target's position in real-time within the drone's system in Fig. \ref{fig:fieldtest} (c) under the \textit{DC} frame. Using Dubbin's curve trajectory planning, the algorithm estimates the target vessel's orientation (Fig. \ref{fig:fieldtest} (d-1)) and approximate length (Fig. \ref{fig:fieldtest} (d-2)) and width (Fig. \ref{fig:fieldtest} (d-3)), providing initial state information for docking and fine-tuning the search. Fig. \ref{fig:fieldtest} (e) shows the heading angle data over the entire duration of the test. The first view of the onshore gimbal camera is listed in Fig. \ref{fig:realtestimag} (a), and drone carrier autonomous approaching and docking to a target vessel in Fig. \ref{fig:realtestimag} (b).

\subsection{Small Object Detection and Grasping}

Once the drone carrier completes its secure attachment, a takeoff command is issued to the UAV, which then ascends into flight (Fig. \ref{fig:realtestimag} (c-1)). Guided by the USV onboard UWB system and utilizing the pre-determined dimensions of the target vessel provided by the USV's radar, the UAV navigates to the centre of the target vessel before initiating a lateral search along its sides. Leveraging its onboard recognition system, preloaded with the target object’s images, the UAV identifies the target, hovers vertically above it (Fig. \ref{fig:realtestimag} (c-2)), and engages its mechanical gripper to attach the target object securely (Fig. \ref{fig:realtestimag} (c-3)). After completing this task, the UAV takes off again, uses UWB positional data to return to the launch point, and performs a precision landing guided by the QR code at the designated location (Fig. \ref{fig:realtestimag} (c-4)).

The path and positional trajectory of the drone during the transportation process are illustrated in Figure 1. In Fig. \ref{fig:dronepath} (a), the drone takes off from the carrier, conducts a search and land on the object. Fig. \ref{fig:dronepath} (b) depicts the drone’s flight with the target and its subsequent landing back on the unmanned carrier. Several key points (P) in these figures represent the takeoff position in the carrier's coordinate system ($P_0$), a position 3 meters above the takeoff point ($P_1$), 3 meters above the target location ($P_2$), and the actual target position ($P_3$), respectively. These positions are measured by the integration of USV onboard UWBs and IMU. However, due to the oscillatory marine environment, IMU errors accumulate throughout the process, leading to significant positional deviations during the drone's return. To ensure a reliable landing, positional information from QR codes is utilized during the taking off and landing phase, effectively enhancing system reliability. 

Maritime testing does not include the operation of a robotic arm or the transportation of large objects using multiple drones; hence, these aspects are not highlighted here. System recovery is achieved by retracting the fixed hook via a winch, releasing the connection between the unmanned vessel and the target ship, and returning to the shore using angular information provided by the onshore gimbal camera.

\section{Conclusion}
This paper presents the design and development of a modular USV-based drone carrier system tailored for maritime inspection and intervention in GNSS-denied environments. The system integrates state-of-the-art sensors, intelligent drones, and a manipulator, embodying a robust, multi-functional robotic architecture. Its fully electric and modular design not only aligns with sustainable operational goals but also enables adaptability to diverse and challenging maritime scenarios. The proposed system has demonstrated significant advancements in autonomous capabilities through extensive experimental validation in both simulated and real-world sea conditions. These trials have confirmed the effectiveness of its perception, recognition, decision-making, and action layers, achieving seamless coordination between the USV and multiple UAVs. The system’s ability to autonomously dock with non-cooperative vessels, perform intervention tasks using manipulators, and transport small targets via drones highlights its potential to address complex maritime challenges with minimal human intervention.


Future research will focus on expanding the carrier’s capabilities, including enhancing multi-drone cooperation, refining manipulator precision, and improving system resilience under extreme environmental conditions. These efforts aim to advance autonomous maritime technologies further, supporting safer and more efficient operations in critical applications.

\section*{Acknowledgment}

The authors would like to express their sincere gratitude to ASPIRE for organizing and sponsoring the MBZIRC2024 competition. Special thanks go to Veena Hausen and Iulia Sisea for their exceptional leadership, coordination, and collaborative efforts, which were instrumental in ensuring the success of this competition.

\bibliographystyle{IEEEtran}
\bibliography{Reference}

\begin{thebibliography}{10}
\providecommand{\url}[1]{#1}
\csname url@samestyle\endcsname
\providecommand{\newblock}{\relax}
\providecommand{\bibinfo}[2]{#2}
\providecommand{\BIBentrySTDinterwordspacing}{\spaceskip=0pt\relax}
\providecommand{\BIBentryALTinterwordstretchfactor}{4}
\providecommand{\BIBentryALTinterwordspacing}{\spaceskip=\fontdimen2\font plus
\BIBentryALTinterwordstretchfactor\fontdimen3\font minus
  \fontdimen4\font\relax}
\providecommand{\BIBforeignlanguage}[2]{{%
\expandafter\ifx\csname l@#1\endcsname\relax
\typeout{** WARNING: IEEEtran.bst: No hyphenation pattern has been}%
\typeout{** loaded for the language `#1'. Using the pattern for}%
\typeout{** the default language instead.}%
\else
\language=\csname l@#1\endcsname
\fi
#2}}
\providecommand{\BIBdecl}{\relax}
\BIBdecl

\bibitem{johnston2017marine}
P.~Johnston and M.~Poole, ``Marine surveillance capabilities of the autonaut
  wave-propelled unmanned surface vessel (usv),'' in \emph{OCEANS
  2017-Aberdeen}.\hskip 1em plus 0.5em minus 0.4em\relax IEEE, 2017, pp. 1--46.

\bibitem{Villa2016}
J.~L. Villa, J.~Paez, C.~Quintero, E.~Yime, and J.~Cabrera, ``Design and
  control of an unmanned surface vehicle for environmental monitoring
  applications,'' in \emph{2016 IEEE Colombian Conference on Robotics and
  Automation (CCRA)}, 2016, pp. 1--5.

\bibitem{jorge2019survey}
V.~Jorge, R.~Granada, R.~Maidana, D.~Jurak, G.~Heck, A.~Negreiros,
  D.~Dos~Santos, L.~Gonçalves, and A.~Amory, ``A survey on unmanned surface
  vehicles for disaster robotics: Main challenges and directions,''
  \emph{Sensors}, vol.~19, no.~3, p. 702, 2019.

\bibitem{matos2013development}
A.~Matos, E.~Silva, N.~Cruz, J.~C. Alves, D.~Almeida, M.~Pinto, A.~Martins,
  J.~Almeida, and D.~Machado, ``Development of an unmanned capsule for
  large-scale maritime search and rescue,'' in \emph{2013 OCEANS-San
  Diego}.\hskip 1em plus 0.5em minus 0.4em\relax IEEE, 2013, pp. 1--8.

\bibitem{Lindemuth2011}
M.~Lindemuth, R.~Murphy, E.~Steimle, W.~Armitage, K.~Dreger, T.~Elliot,
  M.~Hall, D.~Kalyadin, J.~Kramer, M.~Palankar, K.~Pratt, and C.~Griffin, ``Sea
  robot-assisted inspection,'' \emph{IEEE Robotics and Automation Magazine},
  vol.~18, pp. 96--107, 6 2011.

\bibitem{Collins2017}
A.~C. G.~Collins and D.~Twining, ``Enabling technologies for autonomous
  offshore inspections by heterogeneous unmanned teams,'' 2017.

\bibitem{Vasilijevic2015}
A.~Vasilijevic, P.~Calado, F.~Lopez-Castejon, D.~Hayes, N.~Stilinovic, D.~Nad,
  F.~Mandic, P.~Dias, J.~Gomes, J.~C. Molina, A.~Guerrero, J.~Gilabert,
  N.~Miskovic, Z.~Vukic, J.~Sousa, and G.~Georgiou, ``Heterogeneous robotic
  system for underwater oil spill survey,'' in \emph{OCEANS 2015 - Genova},
  2015, pp. 1--7.

\bibitem{Gallego2019}
A.~J. Gallego, A.~Pertusa, P.~Gil, and R.~B. Fisher, ``Detection of bodies in
  maritime rescue operations using unmanned aerial vehicles with multispectral
  cameras,'' \emph{Journal of Field Robotics}, vol.~36, pp. 782--796, 6 2019.

\bibitem{Mendonca2016}
R.~Mendonça, M.~M. Marques, F.~Marques, A.~Lourenço, E.~Pinto, P.~Santana,
  F.~Coito, V.~Lobo, and J.~Barata, ``A cooperative multi-robot team for the
  surveillance of shipwreck survivors at sea,'' in \emph{OCEANS 2016 MTS/IEEE
  Monterey}, 2016, pp. 1--6.

\bibitem{Han2019}
J.~Han, Y.~Cho, and J.~Kim, ``Coastal slam with marine radar for usv operation
  in gps-restricted situations,'' pp. 300--309, 4 2019.

\bibitem{Shen2023}
W.~Shen, Z.~Yang, C.~Yang, and X.~Li, ``A lidar slam-assisted fusion
  positioning method for usvs,'' \emph{Sensors}, vol.~23, 2 2023.

\bibitem{Ma2018}
H.~Ma, E.~Smart, A.~Ahmed, and D.~Brown, ``Radar image-based positioning for
  usv under gps denial environment,'' \emph{IEEE Transactions on Intelligent
  Transportation Systems}, vol.~19, pp. 72--80, 2018.

\bibitem{ferrao2022security}
I.~G. Ferr{\~a}o, D.~Espes, C.~Dezan, and K.~R. L. J.~C. Branco, ``Security and
  safety concerns in air taxis: a systematic literature review,''
  \emph{Sensors}, vol.~22, no.~18, p. 6875, 2022.

\bibitem{jofre2021implementation}
C.~Jofr{\'e}-Brice{\~n}o, F.~Mu{\~n}oz-La~Rivera, E.~Atencio, and R.~F.
  Herrera, ``Implementation of facility management for port infrastructure
  through the use of uavs, photogrammetry and bim,'' \emph{Sensors}, vol.~21,
  no.~19, p. 6686, 2021.

\bibitem{brandao2022side}
A.~S. Brand{\~a}o, D.~Smrcka, {\'E}.~Pairet, T.~Nascimento, and M.~Saska,
  ``Side-pull maneuver: A novel control strategy for dragging a cable-tethered
  load of unknown weight using a uav,'' \emph{IEEE robotics and automation
  letters}, vol.~7, no.~4, pp. 9159--9166, 2022.

\bibitem{wang2023applications}
J.~Wang, K.~Zhou, W.~Xing, H.~Li, and Z.~Yang, ``Applications, evolutions, and
  challenges of drones in maritime transport,'' \emph{Journal of Marine Science
  and Engineering}, vol.~11, no.~11, p. 2056, 2023.

\bibitem{Murphy2008}
R.~R. Murphy, E.~Steimle, C.~Griffin, C.~Cullins, M.~Hall, and K.~Pratt,
  ``Cooperative use of unmanned sea surface and micro aerial vehicles at
  hurricane wilma,'' \emph{Journal of Field Robotics}, vol.~25, pp. 164--180, 3
  2008.

\bibitem{young2017robot}
S.~Young, J.~Peschel, G.~Penny, S.~Thompson, and V.~Srinivasan,
  ``Robot-assisted measurement for hydrologic understanding in data sparse
  regions,'' \emph{Water}, vol.~9, no.~7, p. 494, 2017.

\bibitem{Xiao2017}
X.~Xiao, J.~Dufek, T.~Woodbury, and R.~Murphy, \emph{UAV Assisted USV Visual
  Navigation for Marine Mass Casualty Incident Response}, 2017.

\bibitem{Shao2019}
G.~Shao, Y.~Ma, R.~Malekian, X.~Yan, and Z.~Li, ``A novel cooperative platform
  design for coupled usv-uav systems,'' \emph{IEEE Transactions on Industrial
  Informatics}, vol.~15, pp. 4913--4922, 9 2019.

\bibitem{zhang2020marine}
H.~Zhang, Y.~He, D.~Li, F.~Gu, Q.~Li, M.~Zhang, C.~Di, L.~Chu, B.~Chen, and
  Y.~Hu, ``Marine uav--usv marsupial platform: System and recovery technic
  verification,'' \emph{Applied Sciences}, vol.~10, no.~5, p. 1583, 2020.

\bibitem{Huang2018}
Y.~Huang, Z.~Zheng, L.~Sun, and M.~Zhu, ``Saturated adaptive sliding mode
  control for autonomous vessel landing of a quadrotor,'' \emph{IET Control
  Theory and Applications}, vol.~12, pp. 1830--1842, 9 2018.

\bibitem{li2022synchronized}
W.~Li, Y.~Ge, Z.~Guan, and G.~Ye, ``Synchronized motion-based uav--usv
  cooperative autonomous landing,'' \emph{Journal of Marine Science and
  Engineering}, vol.~10, no.~9, p. 1214, 2022.

\bibitem{wei20223u}
W.~Wei, J.~Wang, Z.~Fang, J.~Chen, Y.~Ren, and Y.~Dong, ``3u: Joint design of
  uav-usv-uuv networks for cooperative target hunting,'' \emph{IEEE
  Transactions on Vehicular Technology}, vol.~72, no.~3, pp. 4085--4090, 2022.

\bibitem{novak2024towards}
F.~Nov{\'a}k, T.~B{\'a}{\v{c}}a, O.~Proch{\'a}zka, and M.~Saska, ``Towards
  uav-usv collaboration in harsh maritime conditions including large waves,''
  \emph{arXiv preprint arXiv:2408.10163}, 2024.

\bibitem{specht2024methodology}
M.~Specht, ``Methodology for performing bathymetric and photogrammetric
  measurements using uav and usv vehicles in the coastal zone,'' \emph{Remote
  Sensing}, vol.~16, no.~17, p. 3328, 2024.

\bibitem{tian2024uav}
E.~Tian, Y.~Li, Y.~Liao, and J.~Cao, ``Uav-usv docking control system based on
  motion compensation deck and attitude prediction,'' \emph{Ocean Engineering},
  vol. 307, p. 118223, 2024.

\bibitem{novak2024collaborative}
F.~Nov{\'a}k, T.~B{\'a}{\v{c}}a, and M.~Saska, ``Collaborative object
  manipulation on the water surface by a uav-usv team using tethers,''
  \emph{arXiv preprint arXiv:2407.08580}, 2024.

\bibitem{Jung2017}
S.~Jung, H.~Cho, D.~Kim, K.~Kim, J.~I. Han, and H.~Myung, ``Development of
  algal bloom removal system using unmanned aerial vehicle and surface
  vehicle,'' \emph{IEEE Access}, vol.~5, pp. 22\,166--22\,176, 10 2017.

\bibitem{Sun2020}
L.~Sun, Y.~Huang, Z.~Zheng, B.~Zhu, and J.~Jiang, ``Adaptive nonlinear relative
  motion control of quadrotors in autonomous shipboard landings,''
  \emph{Journal of the Franklin Institute}, vol. 357, pp. 13\,569--13\,592, 12
  2020.

\bibitem{Liu2023}
X.~Liu, Z.~Hu, Z.~Sun, J.~Lu, W.~Xie, and W.~Zhang, ``A vio-based localization
  approach in gps-denied environments for an unmanned surface vehicle.''\hskip
  1em plus 0.5em minus 0.4em\relax Institute of Electrical and Electronics
  Engineers Inc., 2023, pp. 912--917.

\bibitem{akram2024long}
W.~Akram, S.~Yang, H.~Kuang, X.~He, M.~U. Din, Y.~Dong, D.~Lin, L.~Seneviratne,
  S.~He, and I.~Hussain, ``Long-range vision-based uav-assisted localization
  for unmanned surface vehicles,'' \emph{arXiv preprint arXiv:2408.11429},
  2024.

\bibitem{yolov5}
\BIBentryALTinterwordspacing
G.~Jocher, A.~Chaurasia, A.~Stoken, J.~Borovec, NanoCode012, Y.~Kwon,
  K.~Michael, TaoXie, J.~Fang, imyhxy, Lorna, Z.~Yifu, C.~Wong, A.~V,
  D.~Montes, Z.~Wang, C.~Fati, J.~Nadar, Laughing, UnglvKitDe, V.~Sonck,
  tkianai, yxNONG, P.~Skalski, A.~Hogan, D.~Nair, M.~Strobel, and M.~Jain,
  ``{ultralytics/yolov5: v7.0 - YOLOv5 SOTA Realtime Instance Segmentation},''
  Nov. 2022. [Online]. Available: \url{https://doi.org/10.5281/zenodo.7347926}
\BIBentrySTDinterwordspacing

\bibitem{Zhang2017}
X.~Zhang, W.~Xu, C.~Dong, and J.~M. Dolan, ``{Efficient L-shape fitting for
  vehicle detection using laser scanners},'' \emph{IEEE Intelligent Vehicles
  Symposium, Proceedings}, pp. 54--59, jul 2017.

\bibitem{Cai2023}
W.~Cai, M.~Zhang, Q.~Yang, C.~Wang, and J.~Shi, ``{Long-Range UWB
  Positioning-Based Automatic Docking Trajectory Design for Unmanned Surface
  Vehicle},'' \emph{IEEE Trans. Instrum. Meas.}, vol.~72, 2023.

\bibitem{sun2024aerial}
J.~Sun, Z.~Niu, Y.~Dong, F.~Zhang, M.~U. Din, L.~Seneviratne, D.~Lin,
  I.~Hussain, and S.~He, ``An aerial transport system in marine gnss-denied
  environment,'' \emph{arXiv preprint arXiv:2411.01603}, 2024.

\bibitem{chenheterogeneous}
Q.~Chen, M.~Irfan, Q.~Yu, J.~Sun, M.~U. Din, D.~Lin, S.~He, and I.~Hussain, ``A
  heterogeneous multi-robot system for autonomous object retrieval in
  challenging gnss-denied maritime environment.''

\bibitem{ref:TAD2024}
T.-W. Kang and J.-W. Jung, ``\BIBforeignlanguage{en}{A {Drone}'s {3D}
  {Localization} and {Load} {Mapping} {Based} on {QR} {Codes} for {Load}
  {Management}},'' \emph{\BIBforeignlanguage{en}{Drones}}, vol.~8, no.~4, p.
  130, Mar. 2024.

\bibitem{ref:FRI2004}
F.~Thomas and L.~Ros, ``\BIBforeignlanguage{en}{Revisiting {Trilateration} for
  {Robot} {Localization}},'' \emph{\BIBforeignlanguage{en}{IEEE Transactions on
  Robotics}}, vol.~21, no.~1, pp. 93--101, Feb. 2005.

\bibitem{ref:DKI2020}
D.~Feng, C.~Wang, C.~He, Y.~Zhuang, and X.-G. Xia,
  ``\BIBforeignlanguage{en}{Kalman-{Filter}-{Based} {Integration} of {IMU} and
  {UWB} for {High}-{Accuracy} {Indoor} {Positioning} and {Navigation}},''
  \emph{\BIBforeignlanguage{en}{IEEE Internet of Things Journal}}, vol.~7,
  no.~4, pp. 3133--3146, Apr. 2020.

\bibitem{ref:POF1993}
P.~Grassberger, R.~Hegger, H.~Kantz, C.~Schaffrath, and T.~Schreiber,
  ``\BIBforeignlanguage{en}{On {Noise} {Reduction} {Methods} for {Chaotic}
  {Data}},'' \emph{\BIBforeignlanguage{en}{Chaos: An Interdisciplinary Journal
  of Nonlinear Science}}, vol.~3, no.~2, pp. 127--141, Apr. 1993.

\bibitem{ref:FAC1995}
G.~Welch, G.~Bishop \emph{et~al.}, ``An introduction to the kalman filter,''
  1995.

\bibitem{ref:TSD2019}
T.~Cabreira, L.~Brisolara, and P.~R. Ferreira~Jr.,
  ``\BIBforeignlanguage{en}{Survey on {Coverage} {Path} {Planning} with
  {Unmanned} {Aerial} {Vehicles}},'' \emph{\BIBforeignlanguage{en}{Drones}},
  vol.~3, no.~1, p.~4, Jan. 2019.

\bibitem{ref:TGI2010}
T.~Lee, M.~Leok, and N.~H. McClamroch, ``\BIBforeignlanguage{en}{Geometric
  tracking control of a quadrotor {UAV} on {SE}(3)},'' in
  \emph{\BIBforeignlanguage{en}{49th {IEEE} {Conference} on {Decision} and
  {Control} ({CDC})}}.\hskip 1em plus 0.5em minus 0.4em\relax Atlanta, GA:
  IEEE, Dec. 2010, pp. 5420--5425.

\bibitem{ref:FUM2022}
F.~A.~A. Andrade, I.~P. Guedes, G.~F. Carvalho, A.~R.~L. Zachi, D.~B. Haddad,
  L.~F. Almeida, A.~G. de~Melo, and M.~F. Pinto, ``Unmanned aerial vehicles
  motion control with fuzzy tuning of cascaded-pid gains,'' \emph{Machines},
  vol.~10, no.~1, 2022.

\end{thebibliography}

\begin{IEEEbiography}[{\includegraphics[width=1in,height=1.25in,clip,keepaspectratio]{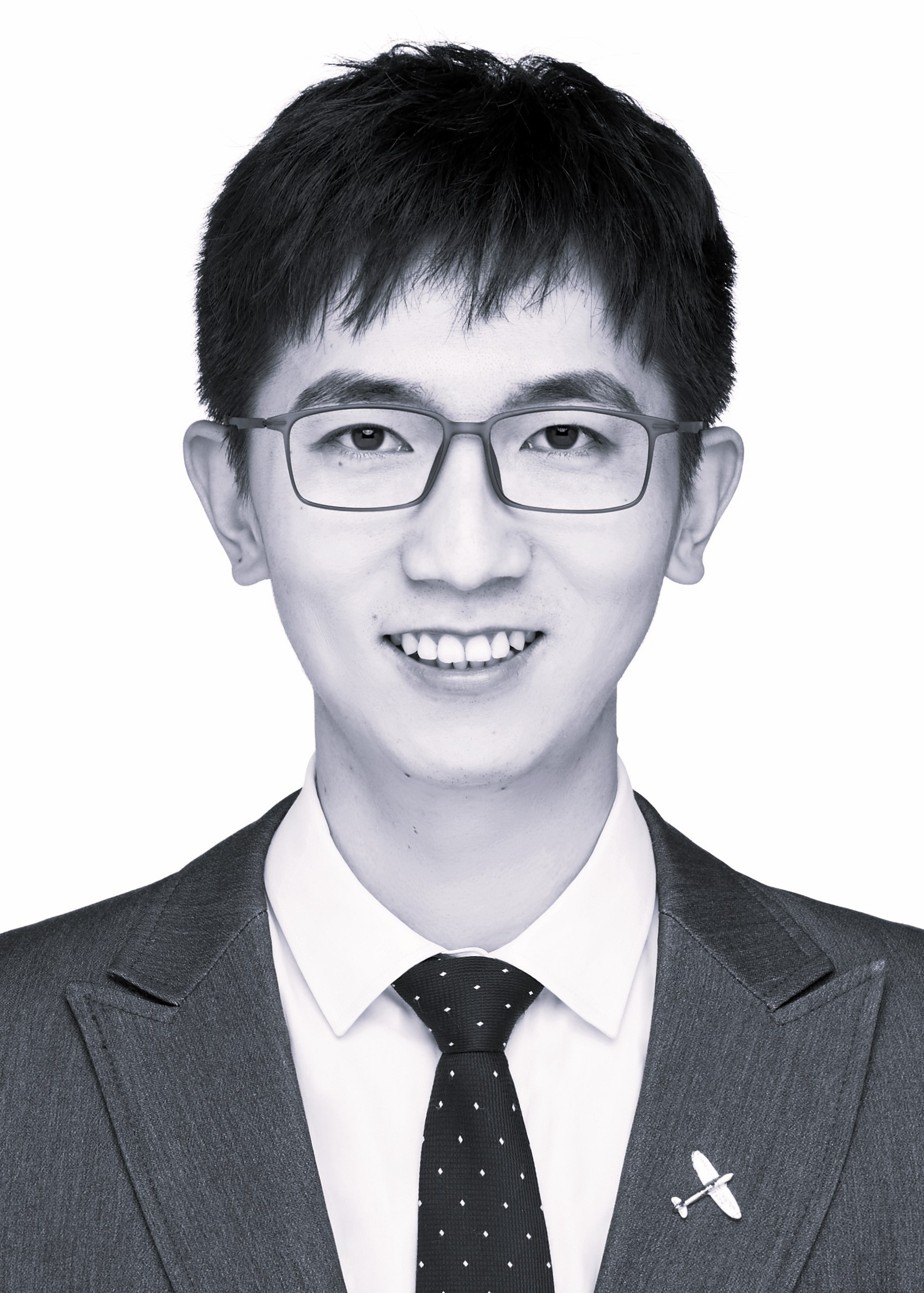}}]{Yihao Dong}
	received the Ph.D. degree in aerospace engineering from Beijing Institute of Technology, Beijing, China, in 2022. He was a post-doctoral researcher at the School of Aeronautic Science and Engineering, Beihang University, Beijing, China. And Visiting Researcher with the Robotics Institute, Khalifa University of Science and Technology, Abu Dhabi. UAE. 
	His research interests include UAV design, intelligent system integration, composite design, and structural topology optimization.
\end{IEEEbiography}

\begin{IEEEbiography}[{\includegraphics[width=1in,height=1.25in,clip,keepaspectratio]{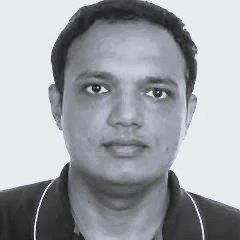}}]{Muhayyu~Ud~Din}
	 received his Ph.D. in Automatic Control, Robotics, and Computer Vision from the Institute of Industrial and Control Engineering at the Universitat Politècnica de Catalunya (UPC), Barcelona, Spain, in 2018. He worked as a Research and Development Engineer at ITK System Engineering from 2019 to 2022, where he was involved in robotic software development. Since 2022, he is working as a Postdoctoral Researcher at the KU centre for Autonomous Robotic Systems (KUCARS) at Khalifa University, Abu Dhabi, UAE. His research interests include marine robotics, motion planning under uncertainties, manipulation planning, and robotic software development.
\end{IEEEbiography}

\begin{IEEEbiography}[{\includegraphics[width=1in,height=1.25in,clip,keepaspectratio]{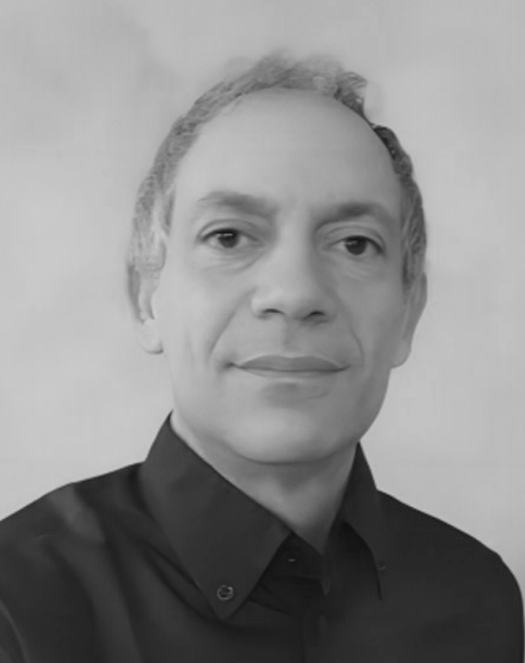}}]{Francesco Lagala}
received a master’s degree in Mechanical Engineering and a Ph. D. in Applied Mechanic from the Department of Automatic Control, University La Sapienza, Italy, in 1994 and 1998, respectively. Since 2000, he has been a researcher at CNR-INM (National Council of Research), where he also served as the director of the experimental setup department from 2005 to 2010. Additionally, La Gala has been an Associate Professor at the University of Roma3 and Sapienza University of Rome since 2003. In 2008, he founded Spinitalia SRL, a private company based in Rome specializing in custom device development. The company has a diverse product portfolio and has been contracted by the Italian army for driverless vehicle applications in recent years. His research interests span electronic, robotic, and mechanical design, as well as computer science, measurement sensors, and fluid dynamics. He has collaborated with leading companies such as Siemens, Airbus, Thales Italia, Leonardo Finmeccanica, and military research departments including the Italian Army, Italian Navy, and Bundeswehr.
\end{IEEEbiography}

\begin{IEEEbiography}[{\includegraphics[width=1in,height=1.25in,clip,keepaspectratio]{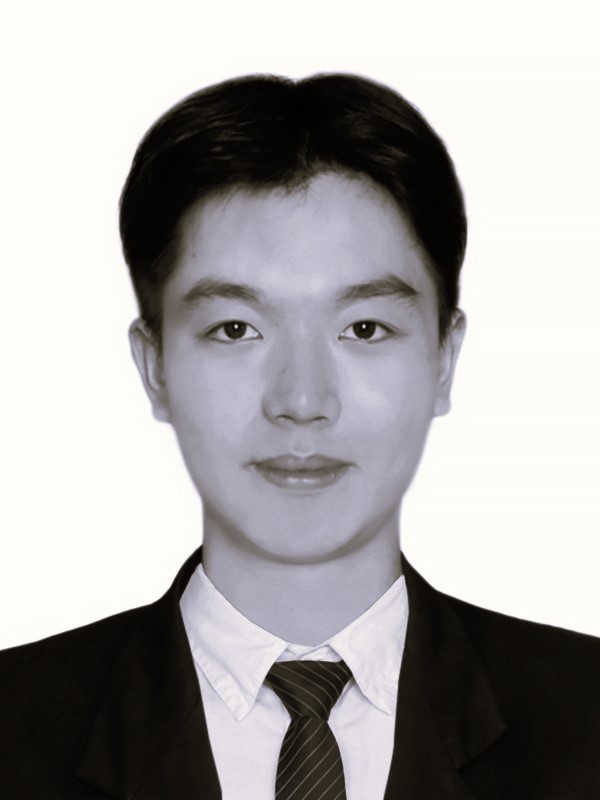}}]{Hailiang Kuang}
	received his B.Sc. degree in Flight Vehicle Design from Beijing Institute of Technology, Beijing, China, in 2021. He is currently working toward the Ph.D. degree in Aerospace Engineering at Beijing Institute of Technology, Beijing, China. His research interests include cooperative control and multi-agent optimization.
\end{IEEEbiography}

\begin{IEEEbiography}[{\includegraphics[width=1in,height=1.25in,clip,keepaspectratio]{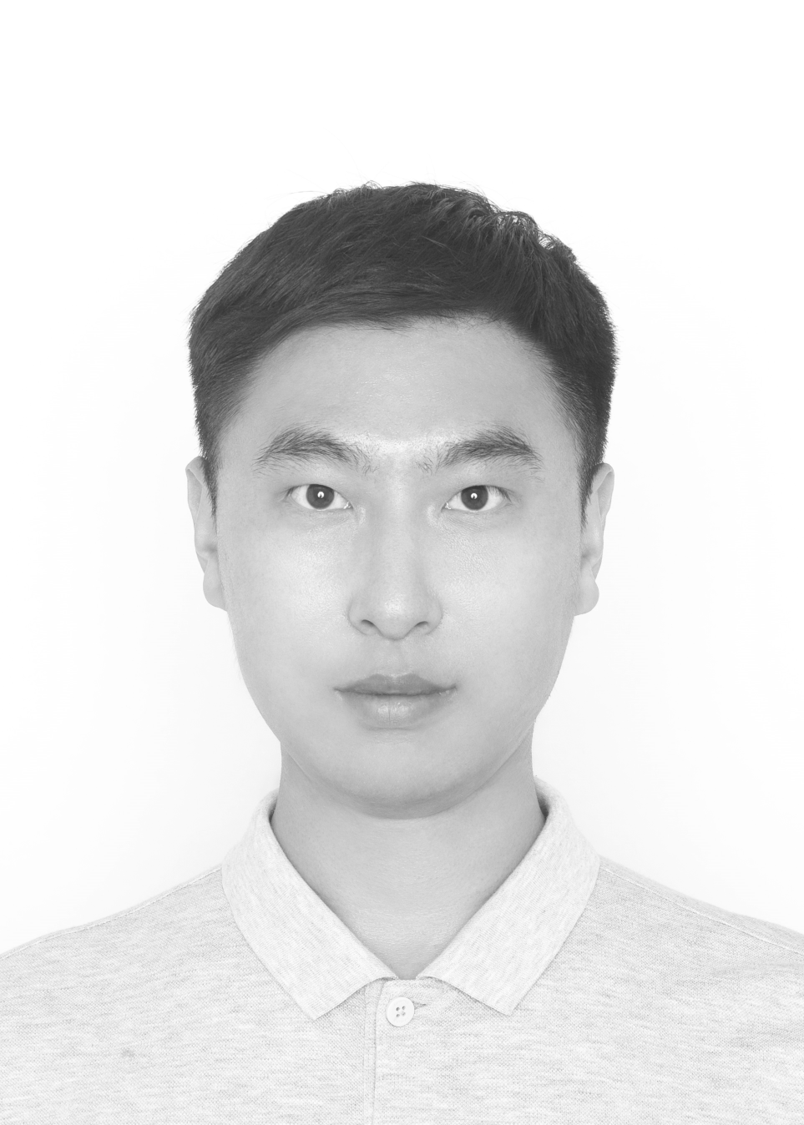}}]{Jianjun Sun}
	received his B.Sc. degree in Information and Computing Science from Taiyuan University of Technology, Taiyuan, China, in 2018, and the M.Sc. degree in control engineering from Beijing Institute of Technology, Beijing, China, in 2021. He is currently working toward the Ph.D. degree in aerospace science and technology at Beijing Institute of Technology, Beijing, China. His research interests include cooperative control, aerospace guidance, and trajectory optimization.
\end{IEEEbiography}

\begin{IEEEbiography}[{\includegraphics[width=1in,height=1.25in,clip,keepaspectratio]{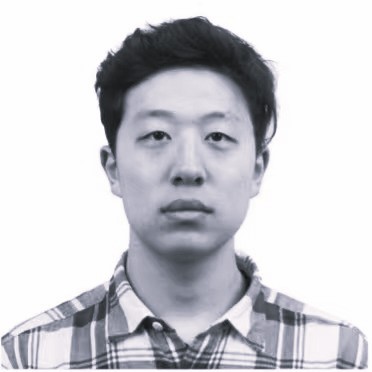}}]{Siyuan Yang}
	 received his B.Sc. degree in Flight Vehicle Design from Beijing Institute of Technology, Beijing, China, in 2021. He is currently working toward the Ph.D. degree in Aerospace Engineering at Beijing Institute of Technology, Beijing, China. His research interests include cooperative control and multi-agent optimization.
\end{IEEEbiography}

\begin{IEEEbiography}[{\includegraphics[width=1in,height=1.25in,clip,keepaspectratio]{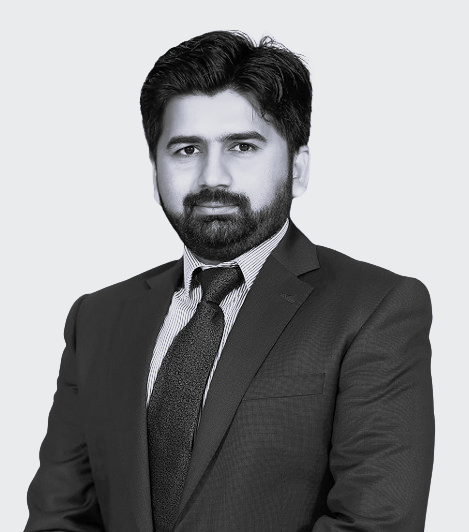}}]{Irfan Hussain}
  received the Ph.D. degree in robotics from the University of Siena, Siena, Italy, in 2016.,He was a Post Doctorate Researcher with the Robotics Institute, Khalifa University of Science and Technology, Abu Dhabi, UAE. He also worked as a Post Doctorate with Siena robotics and system lab (SIRSLab), Siena, Italy. He is currently an Assistant Professor of Robotics and Mechanical Engineering with Khalifa University, Abu Dhabi, UAE. His research interests include embodied intelligence, exoskeletons, extra robotic limbs, soft robotic hands, wearable haptics, grasping and manipulation. Dr. Hussain is the Associate Editor of Proceedings of the Institution of Mechanical Engineers, Part C: Journal of Mechanical Engineering Science and is also the Associate Editor of IEEE/Robotics Automation Society (RAS) International Conference on Robotics and Automation (ICRA)$-$23, ICRA$-$22, ICRA$-$21, RAS$/$IEEE Engineering in Medicine and Biology Society International Conference on Biomedical Robotics, and Biomechatronics $($BioRob$-$18$)$.
\end{IEEEbiography}

\begin{IEEEbiography}[{\includegraphics[width=1in,height=1.25in,clip,keepaspectratio]{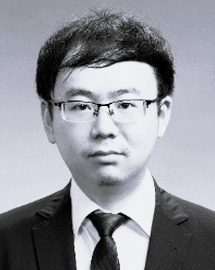}}]{Shaoming He}
	received the B.Sc. and M.Sc. degrees from the Beijing Institute of Technology, Beijing, China, in 2013 and 2016, respectively, and the Ph.D. degree from Cranfield University, Cranfield, U.K., in 2019, all in aerospace engineering. He is currently a professor with the School of Aerospace Engineering, Beijing Institute of Technology, and also a recognized teaching staff with the School of Aerospace, Transport, and Manufacturing, Cranfield University. His research interests include aerospace guidance, multitarget tracking, and trajectory optimization. Dr. He received the Lord Kings Norton Medal award from Cranfield University as the most outstanding doctoral student in 2020.
\end{IEEEbiography}

\end{document}